\documentclass[nohyperref]{article}

\usepackage{url}
\usepackage{algorithm}
\usepackage{algorithmic}
\usepackage{amsthm}
\usepackage{graphicx}
\usepackage{caption}
\usepackage[labelformat=simple]{subcaption}
\usepackage{multirow}
\usepackage{wrapfig}
\usepackage[hidelinks,colorlinks=true,linkcolor=blue,citecolor=cyan]{hyperref}
\usepackage{booktabs}
\usepackage{amssymb}

\newcommand{\xxnote}[3]{}
\ifx\hidenotes\undefined
  \renewcommand{\xxnote}[3]{\color{#2}{#1: #3}}
\fi

\usepackage{microtype}
\usepackage{graphicx}
\usepackage{subcaption}
\usepackage{booktabs} 

\usepackage{hyperref}

\usepackage[accepted]{icml2023}

\usepackage{amsmath}
\usepackage{amssymb}
\usepackage{mathtools}
\usepackage{amsthm}

\usepackage[capitalize,noabbrev]{cleveref}

\theoremstyle{plain}

\theoremstyle{definition}

\theoremstyle{remark}

\usepackage[textsize=tiny]{todonotes}

\icmltitlerunning{A Study of Global and Episodic Bonuses for Exploration in Contextual MDPs}

\begin{document}

\twocolumn[

\icmltitle{A Study of Global and Episodic Bonuses for Exploration in Contextual MDPs}




\begin{icmlauthorlist}
\icmlauthor{Mikael Henaff}{meta}
\icmlauthor{Minqi Jiang}{meta,ucl}
\icmlauthor{Roberta Raileanu}{meta}

\end{icmlauthorlist}

\icmlaffiliation{meta}{Meta AI Research}
\icmlaffiliation{ucl}{University College, London}

\icmlcorrespondingauthor{Mikael Henaff}{mikaelhenaff@meta.com}

\icmlkeywords{Machine Learning, ICML}

\vskip 0.3in
]



\printAffiliationsAndNotice{}  

\begin{abstract}
Exploration in environments which differ across episodes has received increasing attention in recent years. 
Current methods use some combination of \textit{global novelty bonuses}, computed using the agent's entire training experience, and \textit{episodic novelty bonuses}, computed using only experience from the current episode. 
However, the use of these two types of bonuses has been ad-hoc and poorly understood. 
In this work, we shed light on the behavior of these two types of bonuses through controlled experiments on easily interpretable tasks as well as challenging pixel-based settings. 
We find that the two types of bonuses succeed in different settings, with episodic bonuses being most effective when there is little shared structure across episodes and global bonuses being effective when more structure is shared.
We develop a conceptual framework which makes this notion of shared structure precise by considering the variance of the value function across contexts, and which provides a unifying explanation of our empirical results.
We furthermore find that combining the two bonuses can lead to more robust performance across different degrees of shared structure, and investigate different algorithmic choices for defining and combining global and episodic bonuses based on function approximation. 
This results in an algorithm which sets a new state of the art across 16 tasks from the MiniHack suite used in prior work, and also performs robustly on Habitat and Montezuma's Revenge. \phantom{our code is available at web link lalalalala}
\end{abstract}


\section{Introduction}

Balancing exploration and exploitation is a long-standing challenge in reinforcement learning (RL). 
A large body of research has studied this problem within the Markov Decision Process (MDP) framework \citep{sutton&barto}, both from a theoretical standpoint \citep{E3, Rmax, PCPG} and an empirical one. This has led to practical exploration algorithms such as pseudocounts \citep{PseudoCounts}, intrinsic curiosity modules \citep{ICM} and random network distillation \citep{RND}, yielding impressive results on hard exploration problems like Montezuma's Revenge and PitFall~\citep{ALE}.  

More recently, there has been increasing interest in algorithms which move beyond the MDP framework. The standard MDP framework assumes that the agent is initialized in the same environment at each episode (we will refer to these MDPs as \textit{singleton} MDPs). However, several studies have found that agents trained in singleton MDPs exhibit poor generalization, and that even minor changes to the environment can cause substantial degradation in agent performance 
\citep{illuminating_gen, dissection, overfitting_dl, assessing_gen, DBLP:journals/corr/abs-1810-00123, pmlr-v97-cobbe19a, Song2020Observational, DBLP:journals/corr/abs-2111-09794}. 
This has motivated the use of \textit{contextual} MDPs (CMDPs) ~\citep{cmdp}, where different episodes correspond to different environments which nevertheless share structure. Examples of CMDPs include procedurally-generated environments \citep{gym_minigrid, minihack, NLE, obstacle_tower, procgen, deepmindlab, crafter, megaverse} or embodied AI tasks where the agent must generalize across different physical spaces \citep{habitat19iccv, igibson, 3dworld, sapien}.

While exploration is well-studied in the singleton MDP case, it becomes more nuanced when dealing with CMDPs. For singleton MDPs, a common and successful strategy consists of defining an exploration bonus which is added to the reward function being optimized. This exploration bonus typically represents how novel the current state is, where novelty is computed with respect to the entirety of the agent's experience across all episodes. However, it is unclear to what extent this strategy is applicable in the CMDP setting---if two environments corresponding to different episodes are very different, we might not want the experience gathered in one to affect the novelty of a state observed in the other. For example, if an agent is faced with procedurally generated maps with random start and goal locations, exploring the top-left corner of one map does not necessarily mean it should not visit the top-left corner of a different map, since their contents may be different.  

An alternative to using global bonuses is to use episodic ones. Episodic bonuses define novelty with respect to the experience gathered in the current episode alone, rather than across all episodes. Recently, several works \citep{DeepCS, RIDE, AGAC, NovelD, E3B, wang2023revisiting} have used episodic bonuses, with \citet{E3B} and \citet{wang2023revisiting} showing that this is an essential ingredient for solving many sparse reward CMDPs. However, as we will show here, an episodic bonus alone may not be optimal if there is considerable shared structure across different episodes in the CMDP.

In this work, we study the strengths and weaknesses of global and episodic novelty bonuses for exploration in CMDPs, and investigate ways to mitigate their limitations.
First, through a series of easily interpretable examples, 
we show that \textit{global bonuses, which are commonly used in singleton MDPs, can be poorly suited for CMDPs that share little structure across episodes; however, episodic bonuses, which are commonly used in CMDPs, can also fail in cases where knowledge transfer across episodes is crucial.}
We develop a conceptual framework which makes this notion of shared structure precise by considering the variance of the value function in representation space across contexts, providing a unifying explanation of our empirical results. 
Second, we show that by multiplicatively combining episodic and global bonuses, we are able to get more robust performance on both contextual MDPs that share little structure across episodes and singleton MDPs that are identical across episodes. 
We furthermore validate our findings in two challenging pixel-based settings, Habitat \citep{habitat19iccv} and Montezuma's Revenge \citep{ALE}, demonstrating that the tradeoffs between bonus types and advantages of the combined bonus apply there as well.
Third, motivated by these observations, we comprehensively evaluate different combinations of episodic and global bonuses which do not rely on counts, as well as strategies for integrating them, on a wide array of tasks from the MiniHack suite \citep{minihack}.
Our investigations yield an algorithm which combines the elliptical episodic bonus of~\citet{E3B} and the RND global bonus of~\citet{RND}, and sets a new state of the art across $16$ tasks from the MiniHack environment, solving the majority of them. Our code is available at: \url{www.github.com/facebookresearch/e3b}.

\section{Background}

\subsection{Contextual MDPs}

We consider a contextual Markov Decision Process (CMDP) defined by $(\mathcal{S}, \mathcal{A}, \mathcal{C}, P, r, \mu_C, \mu_S)$ where $\mathcal{S}$ is the state space, $\mathcal{A}$ is the action space, $\mathcal{C}$ is the context space, $P$ is the transition function, $\mu_S$ is the initial state distribution conditioned on the context, and $\mu_C$ is the context distribution.
At each episode, we first sample a context $c \sim \mu_C$ and an initial state $s_0 \sim \mu_S(\cdot | c)$. At each step $t$ in the episode, the next state is then sampled according to $s_{t+1} \sim P(\cdot | s_t, a_t, c)$ and the reward is given by $r_t = r(s_t, a_t)$. Let $d_\pi^c$ represent the distribution over states induced by following policy $\pi$ with context $c$. The goal is to learn a policy which maximizes the expected return, averaged across contexts:
\begin{equation*}
R = \mathbb{E}_{c \sim \mu_C, s \sim d_\pi^c, a \sim \pi(\cdot | s)}[r(s, a)]    
\end{equation*}
Examples of CMDPs include procedurally-generated environments, such as ProcGen~\citep{procgen}, MiniGrid \citep{gym_minigrid}, NetHack~\citep{NLE}, or MiniHack~\citep{minihack}, where each context $c$ corresponds to the random seed used to generate the environment. In this case, the number of contexts $|\mathcal{C}|$ is effectively infinite and we will slightly abuse notation by writing $|\mathcal{C}|=\infty$. 
Other examples include embodied AI environments \citep{habitat19iccv, szot2021habitat, 3dworld, igibson, sapien}, where the agent is placed in different simulated houses and must navigate to a location or find an object. 
In this setting, each context $c \in \mathcal{C}$ represents a house identifier and the number of houses $|\mathcal{C}|$ is typically between $20$ and $1000$. 
For an in-depth review of the literature on CMDPs and generalization in RL, see~\cite{gen_review}. Singleton MDPs are a special case of contextual MDPs with  $|\mathcal{C}| = 1$. 

\subsection{Exploration Bonuses}

At a high level, exploration bonuses operate by estimating the novelty of a given state, and assign a high bonus if the state is novel according to some measure. The exploration bonus is then combined with the extrinsic reward provided by the environment, and the result is optimized using RL. More precisely, the reward function optimized by the agent is given by $\bar{r}(s, a) = r(s, a) + \alpha \cdot b(s, a)$, 
where $r(s, a)$ is the extrinsic reward, $b(s, a)$ is the exploration bonus, and $\alpha$ is a parameter governing the balance between exploration and exploitation. Some bonuses do not depend on $a$ or additionally depend the next state $s'$, which will be clear from the context. To account for the variations in scale across different environments and times during training, the exploration bonus is sometimes divided by a running estimate of its standard deviation \cite{RND}. 

In tabular domains with a small number of discrete states, a common choice is to use the inverse counts: $b(s) = 1/\sqrt{N(s)}$ \citep{MBIE}, where $N(s)$ is the number of times state $s$ has been encountered by the agent. However, in most settings of interest the number of states is large or infinite, and many states will not be seen more than once, rendering this  bonus ineffective.  
This has motivated alternative approaches using function approximation. The methods below have proven successful on sparse reward singleton MDPs (RND) and/or sparse reward CMDPs (RIDE, AGAC, NovelD and E3B).

\textbf{Random Network Distillation (RND)} \citep{RND} randomly initializes a neural network $\bar{f}: \mathcal{S} \rightarrow \mathbb{R}^k$, and then trains a second neural network $f$ with the same architecture to predict the outputs of $\bar{f}$ on states encountered by the agent. The exploration bonus associated with a given state $s$ is given by the mean squared error: 
\begin{equation}
b_\mathrm{RND}(s_t) = \|f(s_t) - \bar{f}(s_t)\|_2^2
\end{equation}
The intuition is that for states similar to ones previously encountered by the agent, the error will be low, whereas it will be high for very different states. 
RND has performed well on hard singleton MDPs and is a commonly used component of other exploration algorithms. 

\textbf{Novelty Difference (NovelD)} \citep{NovelD} uses the difference between RND bonuses at two consecutive time steps, regulated by an episodic count-based bonus. Specifically, its bonus is:
\begin{multline}
    b_\mathrm{NovelD}(s_t, a, s_{t+1}) = \\ \Big[b_\mathrm{RND}(s_{t+1}) - c \cdot b_\mathrm{RND}(s_t) \Big]_+ 
    \cdot \mathbb{I}[N_e(s_{t+1}) = 1]    
\end{multline}

Here $b_\mathrm{RND}$ represents the RND bonus defined above, and $N_e(s)$ represents the number of times $s$ has been encountered within the current episode. 
The first term is a \textit{global novelty bonus}, which measures novelty with respect to cross-episode experience, whereas the second term is an \textit{episodic novelty bonus}, which measures novelty with respect to experience within the current episode only. 

\textbf{Adversarially Guided Actor-Critic (AGAC)} \citep{AGAC} also combines global and episodic novelty bonuses. Its bonus is defined by:
\begin{equation}
    b_\mathrm{AGAC}(s_t) = D_\mathrm{KL}(\pi(\cdot | s_t) \| \pi_\mathrm{adv}(\cdot | s_t)) + \beta \frac{1}{\sqrt{N_e(s_t)}}
\end{equation}
where $\pi_\mathrm{adv}$ is a policy trained to mimic the behavior policy $\pi$ (usually with a smaller learning rate). The motivation is that this will encourage the policy to adopt different behaviors as it tries to remain different from the adversary. The second term is an episodic bonus based on $N_e(s)$, the number of times the state $s$ has been encountered within the current episode.

\textbf{Rewarding Impact-Driven Exploration (RIDE)} \citep{RIDE} uses an episodic novelty bonus which is the product of two terms: a count-based reward and the difference between two consecutive state embeddings:
\begin{equation}
    b_\mathrm{RIDE}(s_t) = \frac{1}{\sqrt{N_e(s_t)}} \|\phi(s_{t+1}) - \phi(s_t)\|_2
\label{eq:ride}
\end{equation}

Here the $\phi$ embedding is learned using a combination of inverse and forward dynamics models. The motivation for the second term in the bonus is to reward the agent for taking actions which cause significant changes in the environment. 
RIDE does not use a global novelty bonus.

\textbf{Exploration via Elliptical Episodic Bonuses (E3B)} \citep{E3B} also uses an episodic novelty bonus only, and is motivated by the following observation: while the count-based episodic bonuses used in NovelD, RIDE and AGAC are essential for good performance, they do not scale to complex environments where each state is rarely seen more than once. E3B uses a feature extractor $\phi$ learned using an inverse dynamics model, and defines the episodic bonus as follows:

\begin{align}
    b_\mathrm{E3B}(s_t) &= \phi(s_t)^\top \Big[\sum_{i=t_0}^{t-1} \phi(s_i)\phi(s_i)^\top + \lambda I\Big]^{-1} \phi(s_t)
\label{eq:elliptical_bonus}    
\end{align}

Here $t_0$ denotes the start of the current episode. This can be seen as a generalization of an episodic count-based bonus to continuous state spaces, by noting that it reduces to inverse episodic counts if $\phi$ is a one-hot encoding of the state.

\section{When are Global and Episodic Novelty Bonuses Useful?}
\label{sec:motivation}

Although RIDE, NovelD, AGAC and E3B all use different combinations of episodic and global novelty bonuses, their use in CMDPs has been largely heuristic.
The RIDE and NovelD papers simply state that the episodic bonus is included to prevent the agent from going back and forth between a sequence of states within the same episode.
Furthermore, the global novelty bonuses are justified using the singleton MDP case, but it is unclear to what extent these justifications carry over to the CMDP case. Therefore, a closer investigation of when episodic and global novelty bonuses are useful in CMDPs is required. All experiment details for this section are included in Appendix \ref{appendix:minihack}.

\subsection{Advantages of Episodic Bonuses}
\label{sec:episodic-advantages}

\begin{figure}
     \centering
     \begin{subfigure}[b]{0.235\textwidth}
         \centering
         \includegraphics[width=\textwidth]{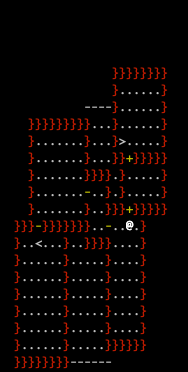}
         \label{fig:y equals x}
     \end{subfigure}
     \hfill
     \begin{subfigure}[b]{0.235\textwidth}
         \centering
         \includegraphics[width=\textwidth]{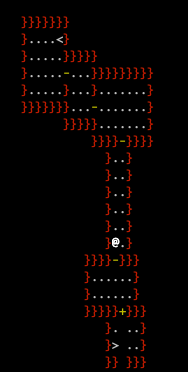}
         \label{fig:three sin x}
     \end{subfigure}
     \hfill
        \caption{Two different contexts of the \texttt{MultiRoom-N6-Lava} environment. Legend: \texttt{\colorbox{black}{\textcolor{white}{@}}}: agent, 
\texttt{\colorbox{black}{\textcolor{white}{<}}}: start, 
\texttt{\colorbox{black}{\textcolor{white}{>}}}: goal, 
\texttt{\colorbox{black}{\textcolor{red}{\}}}}: lava}
        \label{fig:minihack-contexts}
\end{figure}

We begin by providing an example of CMDPs where global novelty bonuses fail and episodic bonuses succeed. Consider the procedurally-generated MiniHack environment shown in Figure \ref{fig:minihack-contexts}. Here, each episode corresponds to a different map where the agent must navigate from the starting location to the goal. The agent only receives reward if it reaches the goal, and the episode terminates if it touches the walls which are made of lava. Because of this, random exploration has a very small chance of reaching the goal before the episode ends, and exploration bonuses are needed. 

We ask the question: are global or episodic novelty bonuses more appropriate here? 
For simplicity, we consider bonuses based on counts of $(x,y)$ locations, which have been commonly used in prior work \citep{AGAC, minihack, NovelD} to avoid the issue of each state being unique:

\begin{equation}
    b_\mathrm{global}(s) = \frac{1}{\sqrt{N(\psi(s))}}, \phantom{a} b_\mathrm{episodic}(s) = \mathbb{I}[N_e(\psi(s)) = 1] \footnote{We also tried $\frac{1}{\sqrt{N_e(\psi(s))}}$, but it performed worse.}
\label{eq:count-based-bonuses}    
\end{equation}

Here $N$ represents counts over all the agent's experience, and $N_e$ represents counts within the current episode only, while $\psi$ is a feature extractor which extracts the $(x, y)$ coordinates of the agent from the state. In general, methods which do not require handcrafted features are preferable, and we focus on them later on in this section and in Section \ref{sec:design-choices}. However, this simple bonus facilitates interpretability, which is the present focus.




We train agents using the global and episodic bonuses in equation \ref{eq:count-based-bonuses} over different numbers of contexts $|\mathcal{C}|$ on the \texttt{MultiRoom} environment shown in Figure \ref{fig:minihack-contexts}. 
The number of contexts represents the number of distinct maps, and one of them is chosen at random at the start of each episode. Results are shown in the top section of Table \ref{tab:count-results}.
The agent using the global bonus consistently obtains near-perfect performance for the singleton MDP setting where $|\mathcal{C}|=1$, but performance steadily degrades as the number of contexts increases. In contrast, when using the episodic bonus, performance remains high as the number of contexts increases, even when $|\mathcal{C}|=\infty$ (no two maps are repeated during training). 
We observed similar trends on two other MiniHack tasks (see Appendix \ref{appendix:extra-minihack-envs}). 
In section \ref{sec:framework}, we provide a framework which explains why the global bonus fails and the episodic bonus is preferable here. 





\subsection{Advantages of Global Bonuses}

\begin{table}[]
    \centering
    \begin{tabular}{l|c|c|c|c}
    Environment & $|\mathcal{C}|$ & $\psi$ & Global & Episodic \\
    \hline
        \texttt{MultiRoom} & $1$ & P & $0.99 \pm 0.00$ & $0.83 \pm 0.23$   \\
        \texttt{MultiRoom} & $3$ & P &$0.59 \pm 0.32$ & $0.92 \pm 0.13$  \\
        \texttt{MultiRoom} & $5$ & P &$0.23 \pm 0.39$ & $0.98 \pm 0.02$  \\         
        \texttt{MultiRoom} & $10$ & P & $0.02 \pm 0.06$ & $0.78 \pm 0.17$  \\         
        \texttt{MultiRoom} & $\infty$ & P & $0.00 \pm 0.00$ & $0.87 \pm 0.10$  \\      
    \hline
        \texttt{Corridors} & $1$ & P & $0.96 \pm 0.03$ & $0.10 \pm 0.68$   \\
    \hline
        \texttt{KeyRoom} & $\infty$ & M & $0.97 \pm 0.00$ & $0.89 \pm 0.01 $ \\
        \texttt{MultiRoom} & $\infty$ & M & $0.99 \pm 0.01$ & $0.59 \pm 0.49$        
    \end{tabular}
    \caption{Reward for global and episodic bonuses for different CMDPs, averaged across $5$ seeds. Performance is close to $0$ for all environments if no bonus is used. Here $|\mathcal{C}|$ denotes the number of different contexts/maps which are sampled from at each episode. The $\psi$ column indicates which feature encodings are used (P for positions, M for messages).}
    \label{tab:count-results}
\end{table}

\begin{figure}
   \begin{center}
     \includegraphics[width=0.3\textwidth]{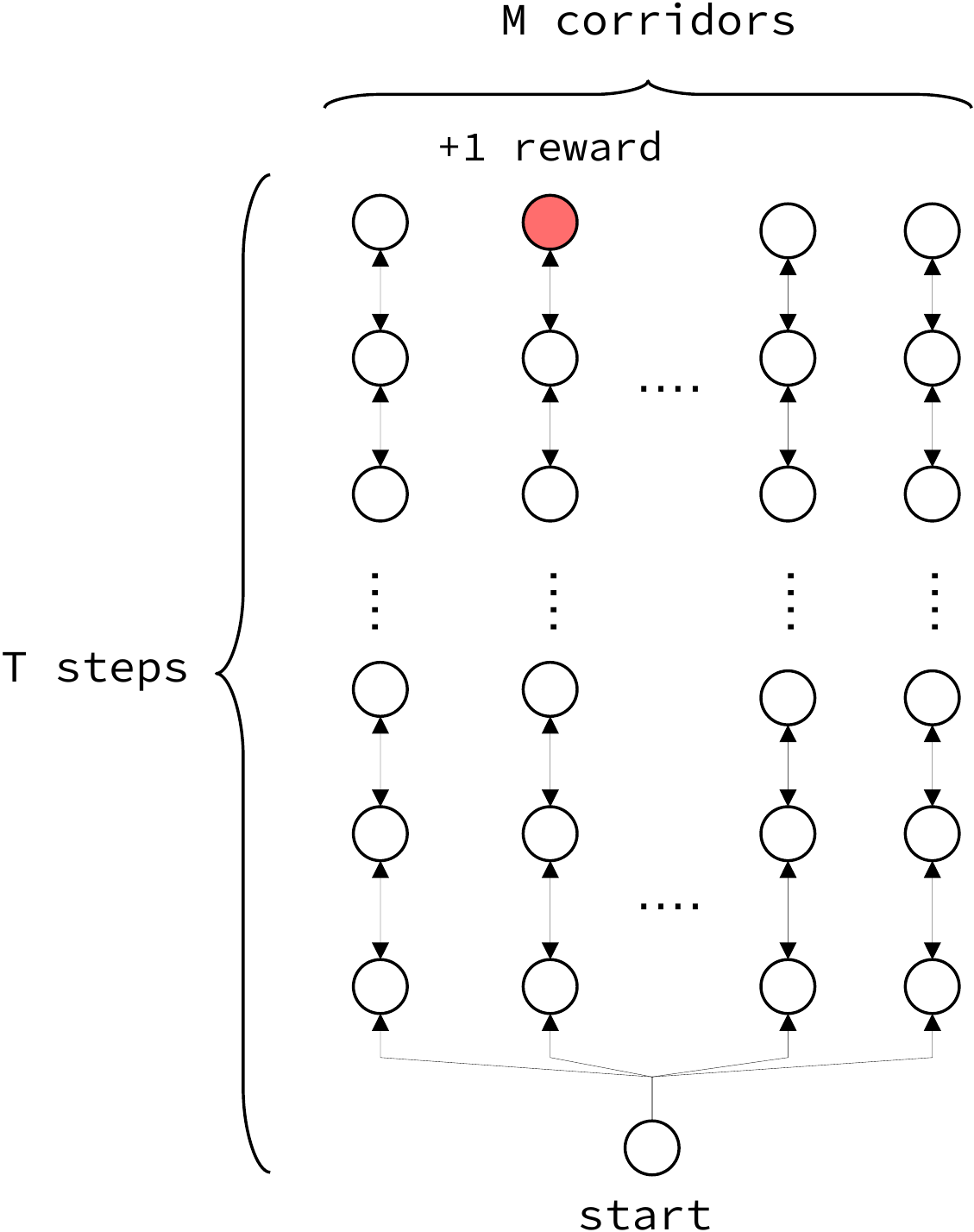}
   \end{center}
   \caption{Simple example where episodic bonus fails.}
   \label{fig:corridors}
 \end{figure}

We next provide an example where the episodic bonus fails but the global bonus succeeds. Consider a singleton MDP with $M$ corridors which can be crossed in $T$ steps, with a single one containing reward at the end (shown in Figure \ref{fig:corridors}). If the episode length is $T$, then any policy which reaches the end of any of the $M$ corridors will get equivalent episodic bonus, and hence the chance of success will be $1/M$. On the other hand, a global bonus will solve the task: after sufficiently visiting one of the corridors, the global bonus there will become depleted and the agent will move on to another one, eventually visiting the corridor with the reward.

\begin{figure}
    \centering
    \includegraphics[width=0.48\textwidth]{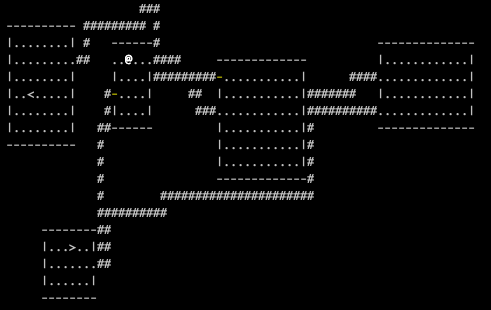}
    \caption{Example map for \texttt{MiniHack-Corridors-R5} enviroment. \texttt{\colorbox{black}{\textcolor{white}{@}}} indicates agent, \texttt{\colorbox{black}{\textcolor{white}{\#}}} corridors connecting rooms, \texttt{\colorbox{black}{\textcolor{white}{<}}} start location and \texttt{\colorbox{black}{\textcolor{white}{>}}} goal.}
    \label{fig:corridors-minihack}
\end{figure}

We illustrate this argument using a singleton version of the \texttt{MiniHack-Corridors-R5} environment (shown in Figure \ref{fig:corridors-minihack}), where the agent must explore different corridors to find its way to the exit. This is similar to the example in Figure \ref{fig:corridors} in the sense that the agent will likely need to explore multiple dead ends before finding the goal. 
Table \ref{tab:count-results} (middle section) shows results for agents trained with the episodic and global bonus on the \texttt{Corridors} environment. In contrast to the previous example, but consistent with our argument above, the global bonus succeeds across all seeds whereas the episodic bonus produces inconsistent performance across seeds, leading to poor performance overall.  

\begin{figure}
    \centering
    \begin{subfigure}[t]{0.45\textwidth}
    \includegraphics[width=0.32\linewidth]{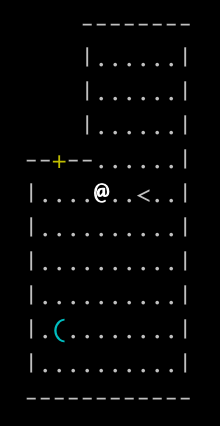}
    \vline
    \includegraphics[width=0.32\linewidth]{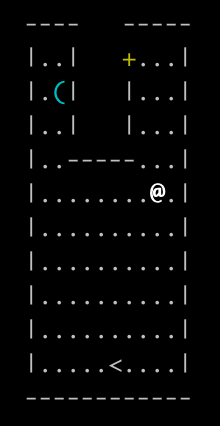}
    \includegraphics[width=0.32\linewidth]{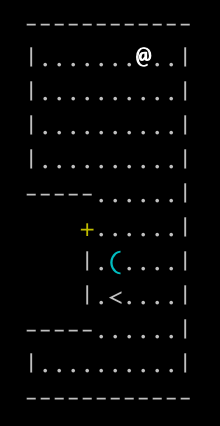}    
    \end{subfigure}%
    \caption{Three example maps for \texttt{MiniHack-KeyRoom-S10} enviroment. \texttt{\colorbox{black}{\textcolor{white}{@}}} indicates agent, \texttt{\colorbox{black}{\textcolor{cyan}{(}}} key, 
     \texttt{\colorbox{black}{\textcolor{yellow}{+}}} door,
     \texttt{\colorbox{black}{\textcolor{white}{<}}} start location and \texttt{\colorbox{black}{\textcolor{white}{>}}} goal. Each observation additionally includes a message such as ``You see here a key of Master Thievery" or ``It's a wall". }
    \label{fig:keyroom-minihack}
\end{figure}

Are global bonuses only useful in the special case of singleton MDPs? We next show that this is not the case, and that they can also be useful in general CMDPs with large $|\mathcal{C}|$. 
We consider the \texttt{KeyRoom} environment, illustrated in Figure \ref{fig:keyroom-minihack}. 
In this environment, the agent must pick up a key and use it to open a door to a small room to reach the exit. Here different contexts correspond to different placements of the agent, key, room, door and exit. We define the $\psi$ feature extractor in equation \ref{eq:count-based-bonuses} to extract the message rather than the $(x, y)$ coordinates (using coordinates does not solve the task for either bonus). We also evaluate both global and episodic bonuses on the \texttt{MultiRoom} environment where $\psi$ extracts messages rather than positions. 
Results in Table \ref{tab:count-results} (bottom section) show that the global bonus solves both environments, even though $|\mathcal{C}|=\infty$ in both cases. 

\subsection{A Framework for Understanding Global and Episodic Novelty Bonuses}
\label{sec:framework}

We now develop a framework for better understanding when and why global and episodic bonuses are effective, and which explains our results so far. Let $\psi: \mathcal{S} \rightarrow Z$ be a feature extractor mapping states to a space $Z$ where novelty bonuses will be computed. This mapping could be hardcoded (as in the examples above) or learned, and states in $\mathcal{S}$ could potentially be high dimensional. 
In Appendix \ref{appendix:value-function-examples}, we show how a number of existing exploration algorithms can be instantiated using this framework, including tabular count-based algorithms, deep RL algorithms that use position or message counts, kernel-based algorithms with global elliptical bonuses, as well as the E3B algorithm which uses a learned $\psi$
 and an episodic bonus.
Now consider the function $V^\star_{\psi, c}: Z \rightarrow [R_{min}, R_{max}]$, defined by:

\begin{equation}
    V^\star_{\psi, c}(z) = \inf_{s \in \psi^{-1}(z)} V^\star(s)
\label{eq:v-phi-c}
\end{equation}

Here $V^\star$ denotes the optimal value function, $R_{min}$ and $R_{max}$ denote the minimum and maximum possible return, and $\psi^{-1}(z) = \{s \in \mathcal{S}_c: \psi(s) = z\}$, where $\mathcal{S}_c$ is the set of states reachable by the agent in context $c$. 
The function $V^\star_{\psi, c}$ can be thought of as a value function over $Z$ corresponding to context $c$.
Note that the infimum ensures that high-value regions in $Z$ correspond to high-value regions in $\mathcal{S}$. We additionally assume that $\psi$ is defined such that there is some subset of $Z$ for which $V^\star_{\psi, c}(z) \approx R_{max}$. This assumption is necessary to rule out pathological cases such as $\psi$ mapping every state to the same point, and holds for all the examples we consider here. 


\textbf{Example 1:} Consider the \texttt{MultiRoom} environment in Figure \ref{fig:minihack-contexts} with positional encodings, i.e. $\psi(s)=(x, y)$, where $(x,y)$ is the location of the agent. Then $Z$ is a 2D lattice the size of the map. The value function $V^\star_{\psi, c}$ will be high centered at the goal and propagate outwards. Note that since the goal changes location for each map, $V^\star_{\psi, c}$ varies significantly across contexts. We verify this empirically and provide visualizations in Figure \ref{fig:value-viz-multiroom} of Appendix \ref{appendix:value-function-viz}. 

\textbf{Example 2:} Consider the \texttt{KeyRoom} environment shown in Figure \ref{fig:keyroom-minihack} with message encodings, i.e. $\psi(s)$ returns the message associated with state $s$. Here $Z$ is the set of all possible messages. The function $V^\star_{\psi, c}(z)$ will then be high for messages indicating that the door has been opened or that the agent has found the key (such as ``You see here a Key of Master Thievery") and low for other messages (``the door is locked"), regardless of the context. Therefore, $V^\star_{\psi, c}$ varies little across contexts.
See Figure \ref{fig:value-viz-keyroom} in Appendix \ref{appendix:value-function-viz} for visualizations.

\textbf{Example 3:} Consider the \texttt{MultiRoom} environment with message encodings. Here $Z$ is the set of all possible messages, and $V^\star_{\psi, c}$ will be high for messages indicating that doors have been opened (such as ``The door opens!"), since this indicates the agent has moved to a new room and is thus closer to the goal. Conversely, $V^\star_{\psi, c}$ will be low for other messages (like the blank message `` ") which do not indicate progress towards the goal. As in the previous example, which messages have high or low values of $V^\star_{\psi, c}$ will not depend much on the context $c$, hence $V^\star_{\psi, c}$ changes little across contexts. 
See Figure \ref{fig:value-viz-multiroom-msg} in Appendix \ref{appendix:value-function-viz} for visualizations. 

\textbf{Example 4:} Consider any singleton MDP, such as the \texttt{Corridors} example from the previous section. Trivially, since the context is always identical, $V^\star_{\psi, c}$ does not change across contexts regardless of $\psi$.

We now argue that global bonuses will fail when $V^\star_{\psi, c}$ changes significantly across different contexts, and succeed when it changes little. To see this, note that a global bonus will induce a sequence of policies $\pi_1, \pi_2, \pi_3, ...$ which progressively visit different parts of the $Z$ space. If $V^\star_{\psi, c}$ varies little across contexts $c$, then eventually some policy $\pi_j$ will visit a part of the $Z$ space which has high value across \textit{all} contexts $c$. Since high value regions in $Z$ correspond to high value regions in $\mathcal{S}$, this means the policy obtains high return across all contexts. 
On the other hand, if $V^\star_{\psi, c}$ varies significantly across contexts, it is more likely that a part of the $Z$ space which was previously visited by policy $\pi_i$ will have high value for some context $c$ which is sampled later on during training. In this case, the agent will no longer visit this region since the global bonus has been exhausted there, thus missing a high-value region in $\mathcal{S}$ as well. 

In contrast to the global bonus, the episodic bonus favors policies which try to cover the \textit{entire} $Z$ space \textit{within each episode}. This is a harder task, and may in fact be impossible if the time horizon is short (see the counterexample in Figure \ref{fig:corridors}). However, if the agent is able to cover the entire $Z$ space within each episode, then they will always visit high-value regions in $Z$ (and thus in $\mathcal{S}$), even if these regions change from one episode to the next---thus avoiding the limitation of the global bonus described above.


This framework provides a consistent explanation for our results so far: recall that the global bonuses succeed in examples 2, 3, 4 (where $V^\star_{\psi, c}$ varies little) and fail in example 1, where $V^\star_{\psi, c}$ varies a lot and the episodic bonus succeeds. 
Note that since $V^\star$ and $\psi$ both appear in the definition of $V^\star_{\psi, c}$ in equation \ref{eq:v-phi-c}, the relative advantage of the global vs. episodic bonuses will depend both on the structure of the CMDP, \textit{and} the feature extractor $\psi$ used to compute the novelty bonus. This framework may also serve to guide practitioners: if sufficient knowledge of the CMDP and $\psi$ is available to estimate how much the $V^\star_{\psi, c}$ function will vary across contexts, this can inform whether to use the global bonus (if it varies little) or the episodic bonus (if it varies a lot). In Appendix \ref{appendix:quantifying-variation} we further discuss how the variation of $V^\star_{\psi, c}$ across episodes can be made precise, and illustrate how it relates to the performance of the global bonus empirically. 



\subsection{Combining Global and Episodic Bonuses}

Our framework described in Section \ref{sec:framework} can provide guidance regarding which bonus to use, when knowledge of the CMDP and feature extractor are available. However, for complex CMDPs or learned feature extractors, it may be difficult to predict how much the $V^\star_{\psi, c}$ function will change across contexts. This motivates the investigation of bonuses which perform robustly across a wide range of CMDPs with differing degrees of shared structure. 

We next investigate a simple strategy whereby we combine global and episodic bonuses via multiplication, which we hypothesize would lead to more robust performance across different regimes compared to either bonus alone.
The resulting combined bonus is given by:

\begin{equation}
    b_\mathrm{combined}(s_t) = \mathbb{I}[N_e(\psi(s_t))=1] \cdot \frac{1}{\sqrt{N(\psi(s_t))}}
    \label{eq:mult-bonus}
\end{equation}

This is motivated by the following observations. First, let us consider the MDP in Figure \ref{fig:corridors}: note that following any of the corridors will maximize the episodic bonus by providing an episodic bonus of $1$ at each step. 
The total combined bonus in equation \ref{eq:mult-bonus} is then equal to the global bonus, and optimizing the global bonus causes the agent to visit each of the corridors until it reaches the one with the reward, solving the MDP.

\begin{table}[]
    \centering
    \begin{tabular}{|l|c|c|c|}
    \hline
    Environment & $\mathcal{C}$ & $\psi$ & Combined Bonus \\
    \hline
         \texttt{MultiRoom} & $\infty$ & P & $0.83 \pm 0.04$  \\
         \texttt{Corridors} & $1$ & P & $0.91 \pm 0.01$ \\
         \texttt{KeyRoom} & $\infty$ & M & $0.99 \pm 0.00$ \\
         \texttt{MultiRoom} & $\infty$ & M & $0.97 \pm 0.00$ \\
    \hline
    \end{tabular}
    \caption{Performance of combined bonus, averaged across $5$ seeds. $|\mathcal{C}|$ denotes number of contexts, P position encodings and M message encodings for $\psi$.}
    \label{tab:combined-results}
\end{table}

Now let us consider the \texttt{MultiRoom} environment with position encodings. If the agent is initialized roughly uniformly throughout the map, the global bonus will decay roughly uniformly across regions over time. This means that the bonus in equation \ref{eq:mult-bonus} will be roughly equal to the episodic bonus (scaled by a constant), which we know is effective.
Finally, in \texttt{KeyRoom} both the episodic and global bonuses will assign high novelty to messages associated with picking up the key, which aligns with the optimal policy, suggesting that their product will also be effective.

Results for all environments are shown in Table \ref{tab:combined-results} (we use the same $\psi$ feature extractor as in previous experiments for each environment).
The combined bonus obtains good performance on all environments, suggesting that it retains the advantages of both the global and episodic bonus here.

\subsection{Scaling to Pixel-Based Settings}

\begin{figure}
     \centering
     \begin{subfigure}[b]{0.23\textwidth}
         \centering
         \includegraphics[width=\textwidth]{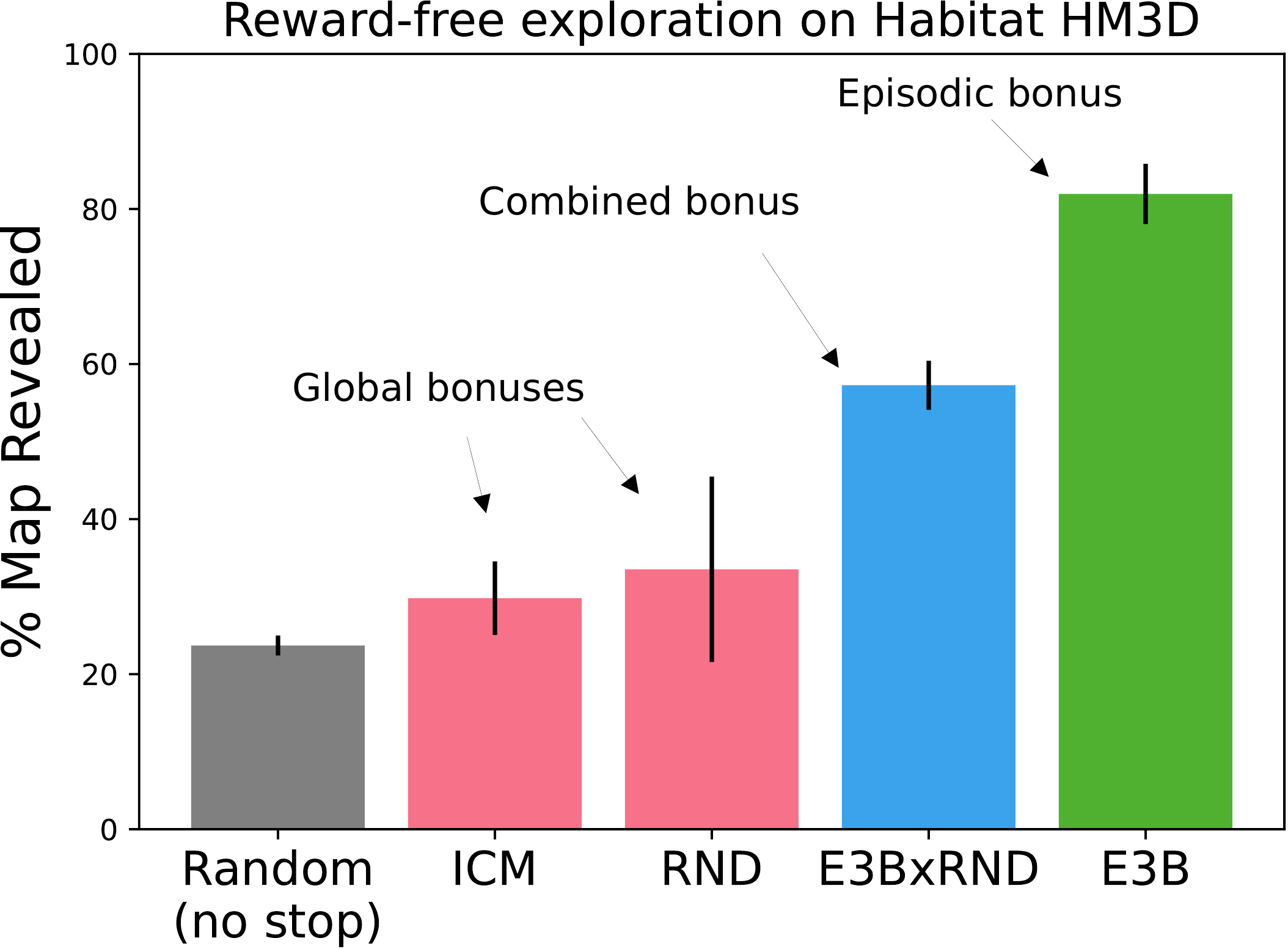}
         \caption{Habitat, $|\mathcal{C}|=1000$}
         \label{fig:habitat-results}
     \end{subfigure}
     \hfill
     \begin{subfigure}[b]{0.23\textwidth}
         \centering
         \includegraphics[width=\textwidth]{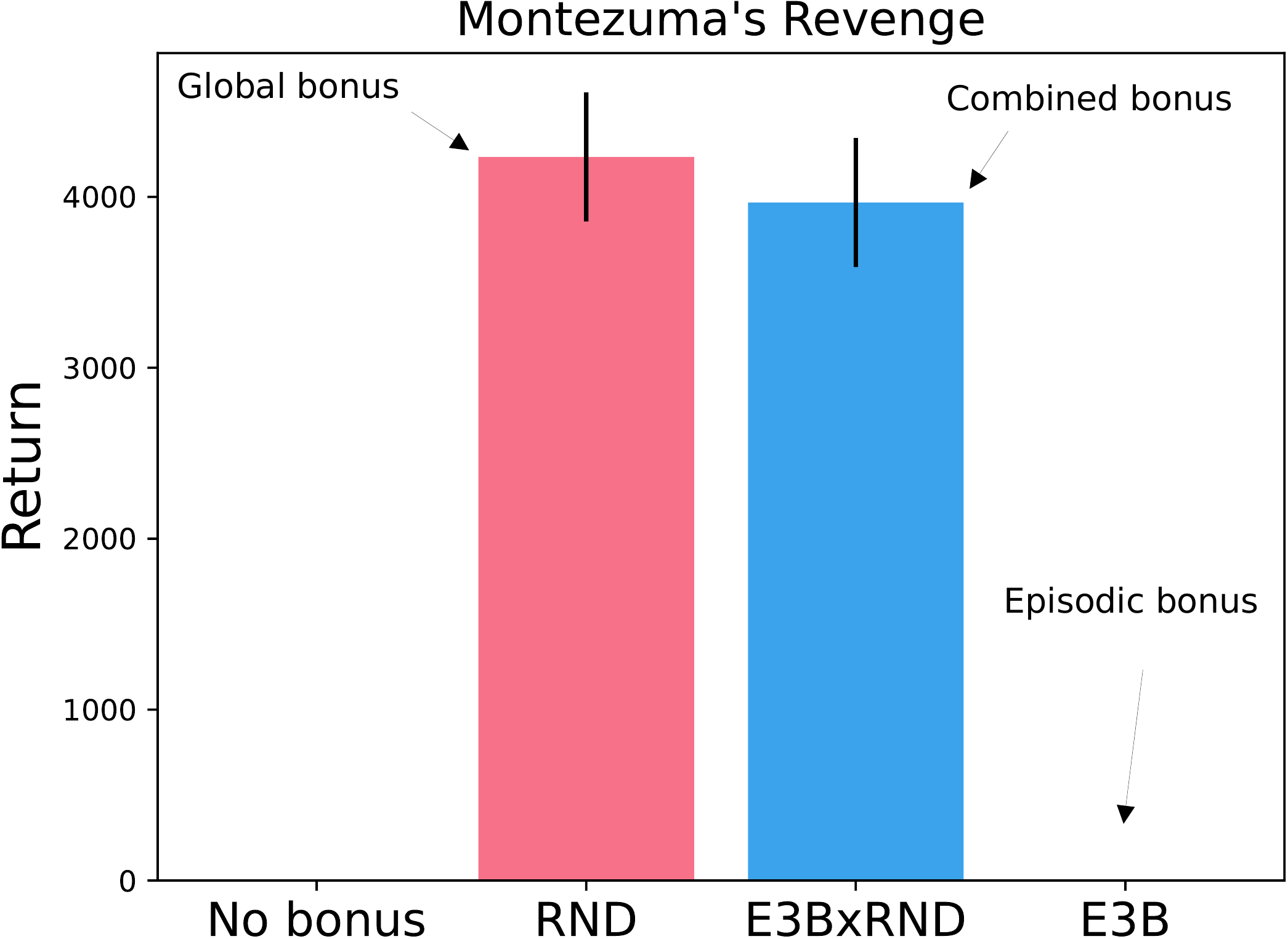}
         \caption{Montezuma, $|\mathcal{C}|=1$}
         \label{fig:montezuma-results}
     \end{subfigure}
     \hfill
        \caption{Comparison of global, episodic and combined bonuses on Habitat and Montezuma's Revenge. Errors bars correspond to standard deviation across $3$ seeds.}
\end{figure}

We next test whether the tradeoffs we have observed between global and episodic bonuses, as well as the advantages of the combined bonus, also apply in high-dimensional, pixel-based settings.
As an example of a pixel-based CMDP with little shared structure across environments, we use Habitat \citep{habitat19iccv}, a photorealistic simulator of indoor environments. Habitat is conceptually similar to the \texttt{MultiRoom} environment in the sense that at each episode, the agent finds itself in a different indoor space consisting of connected rooms. However, the maps in Habitat are considerably more complex and the observations are pixel-based. Here we compare global and episodic bonuses based on function approximation, since counts are not meaningful with high-dimensional images. We use the reward-free exploration setup described in \citet{E3B}, with results shown in Figure \ref{fig:habitat-results}. 
We see that, similarly to \texttt{MultiRoom} with position encodings, the global bonuses (ICM and RND) perform poorly whereas the episodic bonus (E3B) performs well. See Appendix \ref{appendix:habitat} for experiment details. 

We perform a second set of experiments using the Atari environment Montezuma's Revenge \citep{ALE}, a notoriously difficult hard exploration game. This is a singleton MDP where structure is completely shared across contexts, hence our previous results suggest that global bonuses are preferable to episodic ones here. We again compare E3B to RND, with results shown in Figure \ref{fig:montezuma-results}. Consistent with our previous results, we see that the global bonus (RND) performs well, whereas the episodic bonus (E3B) performs poorly. This provides evidence that the tradeoffs we have identified apply more broadly. See Appendix \ref{appendix:montezuma} for experiment details.

For both environments, we also tested a combined bonus obtained by multiplying the episodic bonus from E3B with the global bonus from RND. This approach nearly matches RND's performance on Montezuma's Revenge, and improves upon the global bonus' performance on Habitat, although it does not match the performance of the episodic bonus. 
The combined bonus thus provides more robust performance across scenarios here than an episodic or global bonus alone, although it does not always match the optimal bonus on each task. 

\section{Design Choices for Episodic and Global Novelty Bonuses}
\label{sec:design-choices}

The previous section has shown that global and episodic bonuses succeed in different types of CMDPs, and that combining them via multiplication can yield a bonus which is more robustly effective across tasks. 
However, in order to facilitate interpretability we used count-based bonuses for our MiniHack experiments, which do not scale to complex environments unless task-specific prior knowledge is used (e.g. knowing how to extract $(x,y)$ positions or messages). 
In this section, we perform a thorough a study of global and episodic bonus designs based on function approximation, which do not require such prior knowledge, across a wide range of tasks from the MiniHack suite \citep{minihack}.




\subsection{Experimental Setup}

As our experimental testbed, we use $16$ procedurally-generated tasks from the MiniHack suite \citep{minihack} used in prior work \citep{E3B}. 
The MiniHack tasks are designed to precisely evaluate different capabilities of a given agent, such as navigation, planning or the ability to use objects. Furthermore, many of the MiniHack tasks involve sparse rewards and complex observations which include irrelevant information. 
For evaluation, we follow the protocol suggested by \cite{rliable} and report the mean, median and interquartile mean (IQM) together with $95\%$ confidence intervals using stratified bootstrapping. We use $5$ random seeds for each of the $16$ tasks. Our full experimental details can be found in Appendix \ref{appendix:minihack}.




\subsection{Results}

 We now investigate combining different global novelty bonuses from AGAC, RND and NovelD with the elliptical episodic bonus. We use E3B's elliptical bonus as our episodic bonus instead of a count-based one, since prior work has shown that count-based bonuses either fail in complex environments, or are highly dependent on task-specific feature extractors \citep{E3B}. In contrast, the elliptical bonus has been shown to work well across a wide range of environments without requiring task-specific prior knowledge. We also experimented with the KNN-based episodic bonus from the NGU agent \citep{NGU}, but found that it worked poorly (see Appendix \ref{appendix:ngu-results} for more details). 

Two questions we aim to answer are: i) which global bonus (if any) gives the most improvements when combined with E3B's episodic bonus, and ii) which strategy is best for combining the two bonuses.
To answer this, we consider all possible combinations of E3B's episodic bonus with the global bonuses from RND, NovelD and AGAC, combined either via addition or multiplication. The exact bonuses for each algorithm are detailed in Appendix \ref{appendix:alg-details}. 
We compare to E3B as a baseline since it was previously shown to outperform other methods such as IMPALA, RND, ICM, RIDE and NovelD \citep{E3B}.


\begin{figure}[h]
    \centering
    \includegraphics[width=0.5\textwidth]{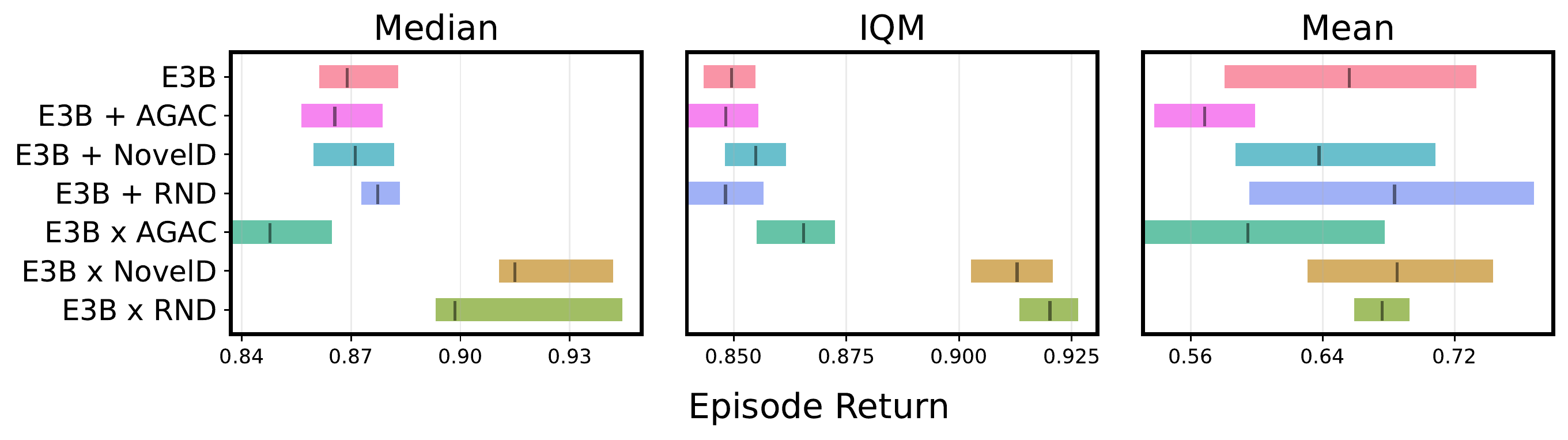}
    \caption{Aggregate performance on $16$ MiniHack tasks. Bars indicate $95\%$ confidence intervals computed using stratified bootstrapping over $5$ seeds.}
    \label{fig:global-bonus-effect}
\end{figure}

Results are shown in Figure \ref{fig:global-bonus-effect}. First, we see that additively combining any of the global bonuses with the elliptical episodic bonus does not provide a meaningful improvement over E3B for any metric. However, multiplicatively combining E3B with either RND or NovelD bonuses produces a large and statistically significant improvement in both median and IQM performance over E3B (the more robust metrics according to~\cite{rliable}). This establishes a new state-of-the-art on MiniHack.

One explanation for the superior performance of the multiplicative combination over the additive one is that the scale of the global bonus decreases significantly throughout training whereas the scale of the episodic bonus does not, since it is reset each episode. Because of this, if we combine the two bonuses via addition, the combined bonus will become increasingly dominated by the episodic bonus. However, if we are combining the two multiplicatively, the global bonus will still have an effect regardless of its scale. See Appendix \ref{appendix:combination-evolution-results} for additional results and discussion.

\section{Related Work}

Exploration in singleton MDPs is a well-studied problem in RL~\citep{sutton&barto, schmidhuber1991possibility, oudeyer2007intrinsic, oudeyer2009intrinsic}. Many theoretical works exist which propose provably efficient algorithms for tabular or linear MDPs \citep{E3, Rmax, MBIE, linearQlearning, OPPO, PCPG, kolter2009near, fruit2017exploration, fruit2018near, fruit2018efficient, tarbouriech2020no}. A number of methods which combine deep RL agents with exploration bonuses have also been proposed for general MDPs~\citep{stadie2015incentivizing, achiam2017surprise}. These include model-free methods such as RND~\citep{RND}, ICM \citep{ICM} and pseudocounts \citep{PseudoCounts, strehl2008analysis, bellemare2016unifying, ostrovski2017count, martin2017count, tang2017exploration, machado2020count}, as well as model-based approaches \citep{max, neural-e3, plan2explore, zhang2021made, manek2021model}. However, these are all designed for the singleton MDP setting and use some form of global bonus which, as we show in Section \ref{sec:motivation}, is not always appropriate to the more general CMDP setting we consider here. We also note the work of \citep{DeepCS}, which showed that episodic bonuses can aid exploration in singleton MDPs. 



More recently, RIDE \citep{RIDE}, AGAC \citep{AGAC}, NovelD \citep{NovelD}, its variants \citep{DBLP:journals/corr/abs-2202-08938}, and others~\citep{parisi2021interesting, campero2020learning, zhang2021made, seurin2021don, fickinger2021explore, tam2022semantic, jo2022leco, ramesh2022exploring} have begun to tackle exploration in procedurally-generated MDPs, a type of CMDP commonly used in empirical research. These methods use combinations of global bonuses designed for singleton MDPs and count-based episodic bonuses. The recent works of~\citet{E3B} and \citet{wang2023revisiting} highlighted the practical importance of episodic bonuses, with \citet{E3B} proposing the elliptical episodic bonus as a solution to the limitations of count-based episodic bonuses, but they did not include a global bonus. Compared to these prior works, our work makes two main contributions. First, whereas previous works justified using global bonuses in CMDPs by appealing to intuitions from singleton MDPs, and provided little justification for using episodic bonuses aside from their empirical performance, we provide clear justifications for the use of each bonus in different types of CMDPs and identify tradeoffs. In particular, we experimentally examine the behavior of each bonus type across different representative settings, and provide a new framework for understanding each one's effect on exploration. This may additionally guide practitioners in choosing an appropriate bonus for the problem at hand.
Second, whereas previous works have investigated different
combinations of global and episodic bonuses in isolation, there has not been a systematic comparison of bonuses and combination strategies, which we perform in Section \ref{sec:design-choices}. This investigation results in a new algorithmic combination which outperforms the previously proposed ones. 

\section{Conclusion}

In this work, we have shed light on the tradeoffs between global and episodic exploration bonuses in CMDPs through experiments in both easily interpretable gridworlds and challenging pixel-based settings, and by developing a new framework which provides a unifying explanation of our empirical results. 
In particular, we find that the effectiveness of each bonus depends on the degree of shared structure between value functions in feature space across different contexts. 
Episodic bonuses tend to be more effective when there is little shared structure across contexts, whereas global bonuses tend to succeed when more structure is shared.
We further show that multiplicatively combining global and episodic bonuses can lead to more robust performance across different settings than either bonus alone. 
Finally, we perform a thorough investigation of design choices for global and episodic bonuses, which leads to an algorithm that sets a new state of the art on a wide range of tasks from the MiniHack suite.  
This work opens up several avenues for future research. Formally quantifying the tradeoffs between global and episodic bonuses through sample complexity bounds presents itself as an intriguing theoretical question. 
Another promising direction is to investigate algorithms which more effectively combine the different bonus types. 
While our multiplicative bonus provides a first step in this direction, it is still limited in the sense that it does not always match the performance of the best bonus type on each individual task. 
More broadly, we hypothesize that agents in rich and ever-changing environments such as NetHack and Minecraft will require both local and global exploration, in order to acquire information at different timescales---how to best achieve this remains an open question. 

\nocite{langley00}

\bibliography{ref}
\bibliographystyle{icml2023}

\newpage
\appendix
\onecolumn

\clearpage

\appendix

\section{Broader Impact Statement}

This work makes progress towards better understanding and designing methods which can efficiently explore contextual MDPs, a very broad framework with applications in video games, virtual reality, autonomous driving, robotics and healthcare. Efficient exploration typically reduces sample complexity, which make real-world application more feasible. Like any RL algorithm, our approach aims to facilitate discovering a policy that maximizes some user-specified reward. Depending on the reward function, executing such a policy could have positive or negative consequences. 

\section{Limitations}

The main technical limitations of this work are twofold. First, although the multiplicative bonus is more robust than the global or episodic bonus in terms of aggregate performance across tasks, it does not always match the best-performing bonus on individual tasks. This is evidenced by our results on Habitat, where the combined bonus still performs worse than the episodic one. We view the multiplicative bonus as a first step towards a method that works across all regimes, but not as a definitive solution. Furthermore, we only consider simple combination strategies for the two bonuses like addition/multiplication, and we do not consider adaptively combining the two based on environment interaction, which is left for future work. Concerning potential negative societal impacts, exploration methods in general can potentially cause harm if deployed in the real world without appropriate safety measures since they seek out novel states, possibly leading to unpredicted behavior. 

\section{Reproducibility Statement}

Our code can be found at: \url{https://github.com/facebookresearch/e3b}. Experiment details can be found in Appendix \ref{appendix:experiments}.

\section{Additional Related Work}
\label{appendix:related}

Exploration in singleton MDPs is a well-studied problem in RL~\citep{sutton&barto, schmidhuber1991possibility, oudeyer2007intrinsic, oudeyer2009intrinsic}. Many theoretical works exist which propose provably efficient algorithms for tabular or linear MDPs \citep{E3, Rmax, MBIE, linearQlearning, OPPO, PCPG, kolter2009near, fruit2017exploration, fruit2018near, fruit2018efficient, tarbouriech2020no}. A number of methods which combine deep RL agents with exploration bonuses have also been proposed for general MDPs~\citep{stadie2015incentivizing, achiam2017surprise}. These include model-free methods such as RND~\citep{RND}, ICM \citep{ICM} and pseudocounts \citep{PseudoCounts, strehl2008analysis, bellemare2016unifying, ostrovski2017count, martin2017count, tang2017exploration, machado2020count}, as well as model-based approaches \citep{max, neural-e3, plan2explore, zhang2021made, manek2021model}. However, these are all designed for the singleton MDP setting and use some form of global bonus which, as we show in Section \ref{sec:motivation}, is not always appropriate to the more general CMDP setting we consider here. We also note the work of \citep{DeepCS} which used episodic bonuses for singleton MDPs. 

\section{Experiment Details}
\label{appendix:experiments}

\subsection{MiniHack}

We used the same architectures and hyperparameters for the experiments with count-based bonuses in Section \ref{sec:motivation} and with function approximation in Section \ref{sec:design-choices}. 

\label{appendix:minihack}
\subsubsection{Architecture Details}

We follow the policy network architecture described in \cite{minihack}. The policy network has four trunks: i) a 5-layer convolutional trunk which maps the full symbol image (of size $79 \times 21$) to a hidden representation, ii) a second 5-layer convolutional trunk which maps a $9 \times 9$ crop centered at the agent to a hidden representation, iii) an MLP trunk which maps the stats vector to a hidden representation, and iv) a 1-D convolutional trunk with interleaved max-pooling layers, followed by a fully-connected network which maps the message to a hidden representation. The hidden representations are then concatenated together, passed through a 2-layer fully-connected network followed by an LSTM \cite{LSTM} layer. The output of the LSTM layer is then passed to linear layers which produce action probabilities and a value function estimate. 

The convolutional trunks i) and ii) have the following hyperparameters: 5 layers, filter size 3, symbol embedding dimension 64, stride 1, filter number 16 at each layer except the last, which is 8, and ELU non-linearities \cite{ELU}. The MLP trunk iii) has 2 hidden layers of 64 hidden units each with ReLU non-linearities.
The trunk iv) for processing messages has 6 convolutional layers, each with 64 input and output feature maps. The first two have kernel size 7 and the rest have kernel size 3. All have stride 1 and there are max-pooling layers (kernel size 3, stride 3) after the 1st, 2nd and 6th convolutional layers. The last two layers are fully-connected and have 128 hidden units and ReLU non-linearities.

For E3B, we used the same architecture as the policy encoder for the feature embedding $\phi$, except we removed the last layers mapping the hidden representation to the actions and value estimate. The inverse dynamics model is a single-layer fully-connected network with 256 hidden units, mapping two concatenated $\phi$ outputs to a softmax distribution over actions.

For RND and NovelD, we also used the same architecture as above for the target and predictor networks. For AGAC, we used the same network as the policy for the adversary.

\subsection{RL Hyperparameters}

For all algorithms we use IMPALA \cite{IMPALA} as our base policy optimizer. Hyperparameters which are common to all methods are shown in Table \ref{tab:hyperparams-common}. All algorithms were trained for 50 million environment steps. We did not anneal learning rates for any of the methods during training. 

Hyperparameters specific to the E3B, RND, NovelD and AGAC components are shown in Tables \ref{tab:hyperparams-e3b}, \ref{tab:hyperparams-rnd} and \ref{tab:hyperparams-noveld}. For all algorithms using an exploration bonus, we used a rolling normalization of the intrinsic reward similar to that proposed in the RND paper \cite{RND}. Specifically, we maintained a running standard deviation $\sigma$ of the intrinsic rewards and divided the intrinsic rewards by $\sigma$ before feeding them to the policy optimizer. For E3B and NovelD, we set the hyperparameters to the values reported in \citep{E3B}. For AGAC, we set the adversary learning rate to be $0.3\times$ the policy learning rate as was done in the official code release. We used the same coefficient for the adversary loss ($0.00004$)--we also experimented with higher values of the adversary loss, but these performed less well.

\begin{table}[h]
  \caption{Common IMPALA Hyperparameters for MiniHack}
  \centering
  \begin{tabular}{ll}
    \toprule
    \hline \\
    Learning Rate & $0.0001$  \\
    RMSProp smoothing constant & $0.99$ \\
    RMSProp momentum & $0$ \\
    RMSProp $\epsilon$ & $10^{-5}$ \\
    Unroll Length     & $80$ \\
    Number of buffers     & $80$       \\
    Number of learner threads & $4$ \\
    Number of actor threads & $256$ \\
    Max gradient norm & $40$ \\
    Entropy Cost & $0.005$ \\
    Baseline Cost & $0.5$ \\
    Discounting Factor & $0.99$ \\
  \end{tabular}
  \label{tab:hyperparams-common}
\end{table}

\begin{table}[h]
  \caption{E3B Hyperparameters}
  \centering
  \begin{tabular}{ll}
    \toprule
    \hline \\
    Ridge $\lambda$ & $0.1$  \\
    Intrinsic Reward Normalization & True \\
    Intrinsic Reward Coefficient & $1.0$ \\
  \end{tabular}
  \label{tab:hyperparams-e3b}
\end{table}

\begin{table}[h]
  \caption{RND Hyperparameters}
  \centering
  \begin{tabular}{ll}
    \toprule
    \hline \\
    Predictor Learning Rate & $0.0001$  \\
    Intrinsic Reward Normalization & True \\
    Intrinsic Reward Coefficient & $1.0$ \\
  \end{tabular}
  \label{tab:hyperparams-rnd}
\end{table}

\begin{table}[h]
  \caption{NovelD Hyperparameters}
  \centering
  \begin{tabular}{ll}
    \toprule
    \hline \\
    Predictor Learning Rate & $0.0001$  \\
    Scaling Factor $c$ & $0.1$ \\
    Intrinsic Reward Normalization & True \\
    Intrinsic Reward Coefficient & $1.0$ \\
  \end{tabular}
  \label{tab:hyperparams-noveld}
\end{table}

\begin{table}[h]
  \caption{AGAC Hyperparameters}
  \centering
  \begin{tabular}{ll}
    \toprule
    \hline \\
    Adversary Learning Rate & $0.00003$  \\
    Adversary loss term & $0.00004$ \\
    Intrinsic Reward Normalization & True \\
    Intrinsic Reward Coefficient & $1.0$ \\
  \end{tabular}
  \label{tab:hyperparams-agac}
\end{table}

For the experiments in Section 4.2, we tuned the $\beta$ hyperparameter over the range $\{1, 10, 100, 1000, 10000, 100000\}$. In initial experiments we noticed that the episodic bonus was several orders of magnitude smaller than the episodic bonus, hence we used a high range of values for the $\beta$ hyperparameter to bring the global bonus to a similar range.

\clearpage
\subsection{Task Details}
\label{appendix:task-details}

We used the following set of $16$ MiniHack tasks:
\texttt{ 'MiniHack-MultiRoom-N4-Locked-v0',
         'MiniHack-MultiRoom-N6-Lava-v0',
         'MiniHack-MultiRoom-N6-Lava-OpenDoor-v0',
         'MiniHack-MultiRoom-N6-LavaMonsters-v0',
         'MiniHack-MultiRoom-N10-v0', 
         'MiniHack-MultiRoom-N10-OpenDoor-v0', 
         'MiniHack-MultiRoom-N10-Lava-v0', 
         'MiniHack-MultiRoom-N10-Lava-OpenDoor-v0', 
         'MiniHack-LavaCrossingS19N13-v0',
         'MiniHack-LavaCrossingS19N17-v0',             
         'MiniHack-Labyrinth-Big-v0',
         'MiniHack-Levitate-Potion-Restricted-v0',
         'MiniHack-Levitate-Boots-Restricted-v0',
         'MiniHack-Freeze-Horn-Restricted-v0', 
         'MiniHack-Freeze-Wand-Restricted-v0', 
         'MiniHack-Freeze-Random-Restricted-v0'}. 

Note that the \texttt{-Restricted-} versions of the tasks have restricted action spaces, as described in \citep{E3B}.

\subsection{Habitat}
\label{appendix:habitat}

\subsubsection{Environment Details}

\begin{figure}[h]
     \centering
     \begin{subfigure}[b]{0.6\textwidth}
         \centering
         \includegraphics[width=\textwidth]{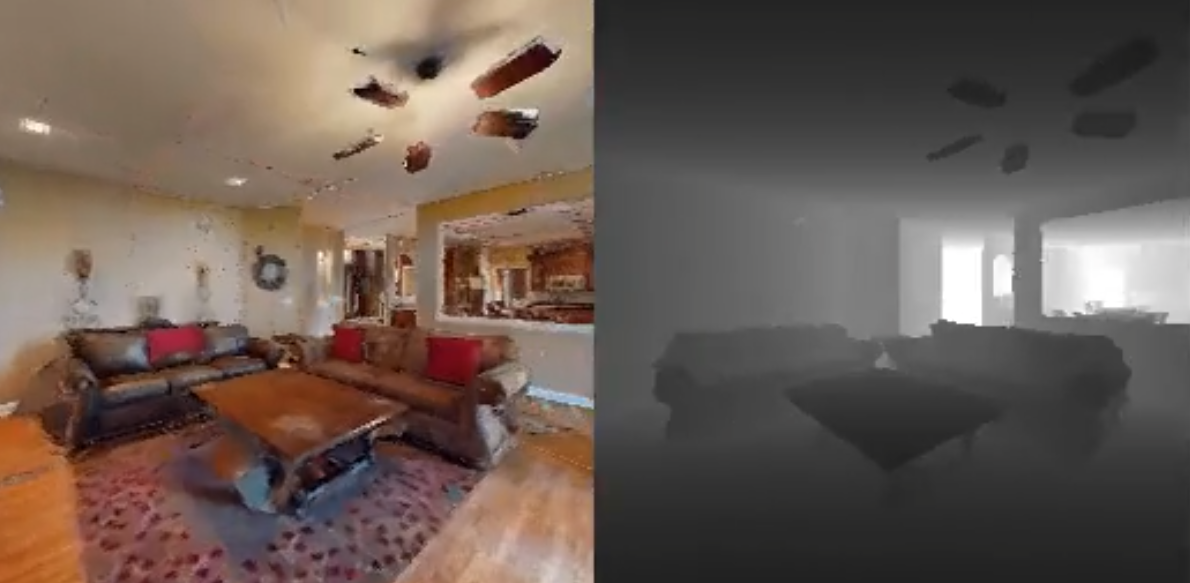}
         \caption{}
     \end{subfigure}
     \hfill
     \begin{subfigure}[b]{0.3\textwidth}
         \centering
         \includegraphics[width=\textwidth]{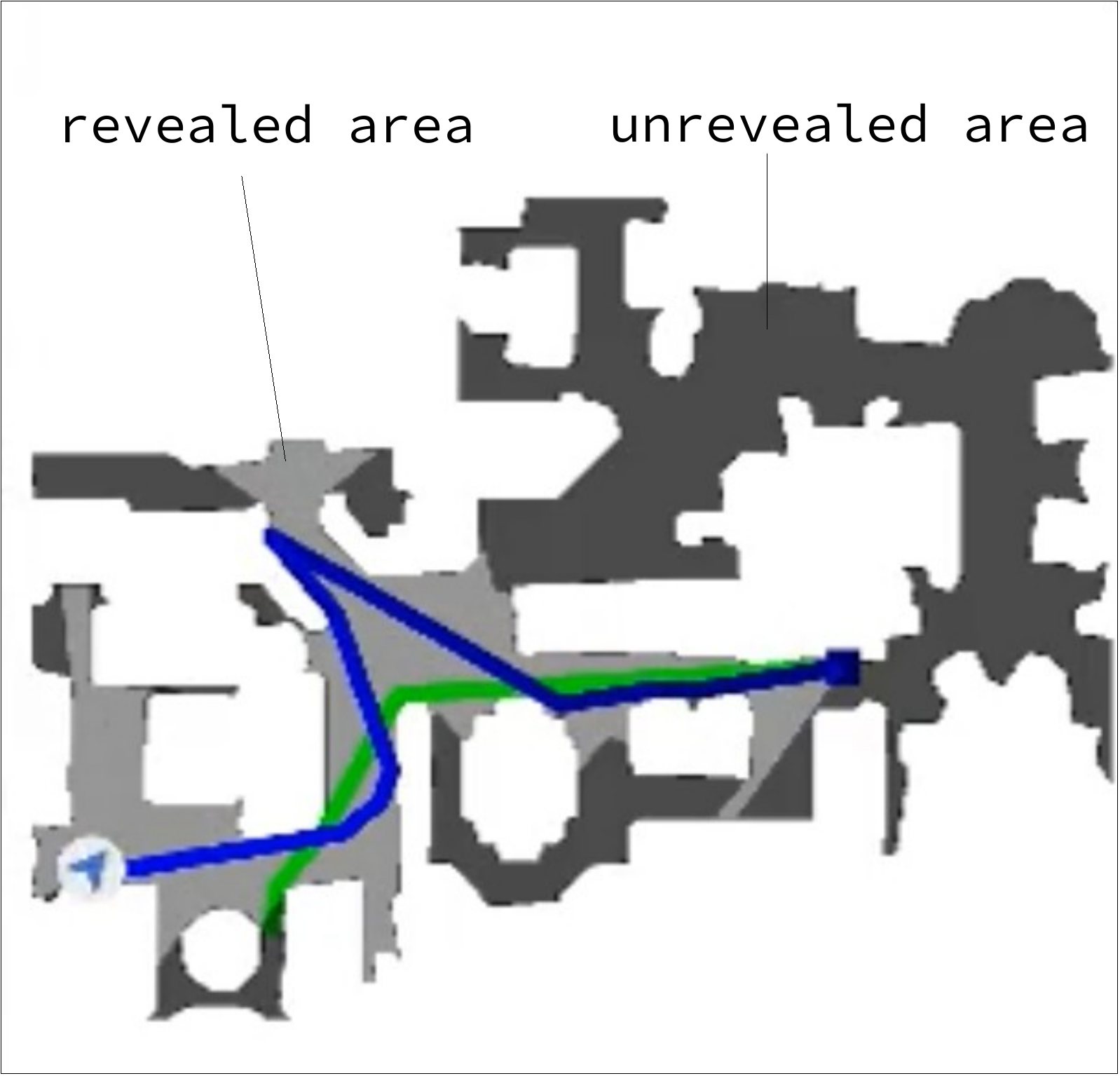}
         \caption{}
     \end{subfigure}
     \caption{a) Visual observations in Habitat  b) Exploration is measured as the proportion of the environment revealed by the agent's line of sight over the course of the episode.}
     \label{fig:habitat2}
\end{figure}

We used the HM3D \cite{hm3d} dataset, which consists of $1000$ high-quality renderings of indoor scenes. 
Observations consist of $4$ modalities: an RGB and depth image (shown in Figure \ref{fig:habitat2}a), GPS coordinates and the compass heading. The action space consists of $4$ actions: $\mathcal{A}$ = \texttt{\{stop\_episode, move\_forward (0.25m), turn\_left ($10^\circ$), turn\_right ($10^\circ$)\}}. The dataset scenes are split into $800/100/100$ train/validation/test splits. Since the test split is not publicly available, we evaluate all models on the validation split. Each scene corresponds to a different context $c \in \mathcal{C}$ in the CMDP framework. 

To measure exploration coverage, we compute the area revealed by the agent's line of site using the function provided by the Habitat codebase \footnote{\url{https://github.com/facebookresearch/habitat-lab/blob/main/habitat/utils/visualizations/fog_of_war.py}}, which uses a modified version of Bresenham's line cover algorithm. We define the exploration coverage to be:

\begin{equation*}
    \mbox{coverage} = \frac{\mbox{revealed area}}{\mbox{total area}}
\end{equation*}

See Figure \ref{fig:habitat2}b) for an illustration. For the results in Figure \ref{fig:habitat-results}, we evaluated exploration performance for each algorithm by measuring its coverage on $100$ episodes using scenes from the validation set (which were not used for training). 

\subsubsection{Architecture Details}

For all Habitat experiments we used the same policy network as in \cite{DDPPO}, which includes a ResNet50 visual encoder \cite{resnet} and a 2-layer LSTM \cite{LSTM} policy. In addition to RGB and Depth images, the agent also receives GPS coordinates and compass orientation, represented by $3$ scalars total, which are fed into the policy. See the official code release at \url{https://github.com/facebookresearch/habitat-lab/tree/main/habitat_baselines} for full details. 

For exploration algorithms which use inverse dynamics models (E3B and ICM), we set the architecture of the encoder $\phi$ to be identical to that of the policy network, except that the last layer mapping hidden units to actions is removed. The inverse dynamics model was a single layer MLP with $256$ hidden units and ReLU non-linearities. 

For exploration algorithms which use random network distillation (RND and NovelD), we set the architecture of the random network to be identical to that of the policy network.

\subsubsection{RL Hyperparameters}

The DD-PPO hyperparameters which are common to all the algorithms are listed in Table \ref{tab:hyperparams-common-habitat}. The hyperparameters which are specific to each algorithm are listed in Table \ref{tab:hyperparams-habitat-e3b}, \ref{tab:hyperparams-habitat-rnd}, \ref{tab:hyperparams-noveld}. For NovelD's count-based bonus, hashing the full image was too slow to be practical, so we subsampled images by a factor of $1000$ used that for the count-based bonus, along with the GPS coordinates and compass direction.

\begin{table}[h!]
  \caption{Common PPO/DD-PPO Hyperparameters for Habitat}
  \centering
  \begin{tabular}{ll}
    \toprule
    \midrule
    Clipping & $0.2$  \\    
    PPO epochs & $2$  \\    
    Number of minibatches & $2$ \\
    Value loss coefficient & $0.5$ \\
    Entropy coefficient & 0.00005 \\
    Learning rate & $0.00025$ \\
    $\epsilon$ & $10^{-5}$ \\
    Max gradient norm & $0.2$ \\
    Rollout steps & $128$ \\
    Use GAE & True \\
    $\gamma$ & 0.99 \\
    $\tau$ & 0.95 \\
    Use linear clip decay & False \\
    Use linear LR decay & False \\
    Use normalized advantage & False \\
    Hidden size & $512$ \\
    DD-PPO Sync fraction & $0.6$ \\
    \bottomrule
  \end{tabular}
  \label{tab:hyperparams-common-habitat}
\end{table}

\begin{table}[h!]
  \caption{Hyperparameters for E3B on Habitat}
  \centering
  \begin{tabular}{|l|l|l|}
    \hline
    Hyperparameter & Values considered & Final Value \\
    \hline
    Ridge regularizer $\lambda$ & $\{0.1\}$ & $0.1$  \\
    Intrinsic reward coefficient $\beta$ & $\{1.0, 0.1, 0.01, 0.001, 0.0001$ & $0.1$ \\
    Inverse Dynamics Model updates per PPO epoch & $3$ & $3$ \\
    \hline
  \end{tabular}
  \label{tab:hyperparams-habitat-e3b}
  \vspace{2mm}
  \caption{Hyperparameters for RND on Habitat}
  \begin{tabular}{|l|l|l|}
    \hline
    Hyperparameter & Values considered & Final Value \\  
  \hline
    Intrinsic reward coefficient $\beta$ & $\{1.0, 0.1, 0.01, 0.001, 0.0001$ & $0.1$ \\
    Predictor Model updates per PPO epoch & $3$ & $3$ \\
    \hline
  \end{tabular}  
  \label{tab:hyperparams-habitat-rnd}
  \caption{Hyperparameters for ICM on Habitat}
  \begin{tabular}{|l|l|l|}
    \hline
    Hyperparameter & Values considered & Final Value \\
    \hline
    Intrinsic reward coefficient $\beta$ & $\{1.0, 0.1, 0.01, 0.001, 0.0001$ & $0.1$ \\
    Forward Dynamics Model loss coefficient & $\{1.0\}$ & $1.0$ \\
    \hline
  \end{tabular}  
  \label{tab:hyperparams-habitat-icm}

\end{table}

\subsubsection{Compute Details}

Each job was run for $225$ million steps, which took approximately $3$ days on $32$ GPUs with $10$ CPU threads.

\clearpage

\subsection{Experiment Details for Montezuma's Revenge}
\label{appendix:montezuma}

We used the PyTorch RND implementation from \url{https://github.com/jcwleo/random-network-distillation-pytorch} as our base codebase, which reimplements \citet{RND}. The policy network consists of $3$ convolutional layers with $32/64/64$ feature maps followed by an MLP with $3$ hidden layers with $256/448/448$ hidden units respectively and ReLU non-linearities. The RND predictor network consists of $3$ convolutional layers with $32/64/64$ feature maps and Leaky ReLUs followed by a 2-layer MLP with $512$ hidden units and standard ReLUs. The target network is the same as the predictor network, except it only has a single hidden layer MLP following the convolutional layers.

For E3B we used the same network architecture as for the predictor network for the $\psi$ encoder. We experimented with both learning the $\psi$ encoder using an inverse dynamics model and using a fixed random network, which has been shown to work well in certain cases \citep{pathak18largescale}, and found that using a fixed random network worked best. We tuned the ridge regularization for the covariance matrix over the range $\{0.01, 0.1, 1.0\}$ and kept $1.0$ as the final value.

\begin{table}[h!]
  \caption{Hyperparameters for PPO+RND on Montezuma's Revenge}
  \centering
  \begin{tabular}{|l|l|l|}
    \hline
    Hyperparameter & Value \\
    \hline
Max Step Per Episode & 4500 \\
Extrinsic Reward Coefficient & 2 \\
Intrinsic Reward Coefficient & 1. \\
Learning Rate & 1e-4 \\
Num. Env & 128 \\
Rollout length & 128 \\
$\gamma$ & 0.999 \\
Intrinsic $\gamma$ & 0.99 \\
$\lambda$ & 0.95 \\
StableEps & 1e-8 \\
Frame Stack & 4 \\
Image Height & 84 \\
Image Width & 84 \\
UseGAE & True \\
Gradient Clipping Norm & 0.5 \\
Entropy & 0.001 \\
Epoch & 4 \\
MiniBatch & 4 \\
PPOEps & 0.1 \\
ActionProb & 0.25 \\
UpdateProportion & 0.25 \\
ObsNormStep & 50 \\
    \hline
  \end{tabular}
\end{table}

\clearpage

\section{Algorithm Details for MiniHack Experiments}
\label{appendix:alg-details}

The bonuses for each algorithm we consider are detailed below:

\begin{align*}
    b_\mathrm{E3B \times AGAC}(s_t) &= \Big[\phi(s_t)^\top C_{t-1}^{-1}\phi(s_t)\Big] \cdot D_\mathrm{KL}(\pi(\cdot | s_t) \| \pi_\mathrm{adv}(\cdot | s_t)) \\
    b_\mathrm{E3B \times RND}(s_t) &= \Big[\phi(s_t)^\top C_{t-1}^{-1}\phi(s_t)\Big] \cdot \|f(s_t) - \bar{f}(s_t)\|_2^2 \\
    b_\mathrm{E3B \times NovelD}(s_t) &= \Big[\phi(s_t)^\top C_{t-1}^{-1}\phi(s_t)\Big] \cdot \Big[\|f(s_{t+1}) - \bar{f}(s_{t+1})\|_2^2 - c\|f(s_t) - \bar{f}(s_t)\|_2^2\Big]_+ \\
    b_\mathrm{E3B + AGAC}(s_t) &= \Big[\phi(s_t)^\top C_{t-1}^{-1}\phi(s_t)\Big] + \beta D_\mathrm{KL}(\pi(\cdot | s_t) \| \pi_\mathrm{adv}(\cdot | s_t)) \\
    b_\mathrm{E3B + RND}(s_t) &= \Big[\phi(s_t)^\top C_{t-1}^{-1}\phi(s_t)\Big] + \beta \|f(s_t) - \bar{f}(s_t)\|_2^2 \\
    b_\mathrm{E3B + NovelD}(s_t) &= \Big[\phi(s_t)^\top C_{t-1}^{-1}\phi(s_t)\Big] + \beta \Big[\|f(s_{t+1}) - \bar{f}(s_{t+1})\|_2^2 - c\|f(s_t) - \bar{f}(s_t)\|_2^2\Big]_+    
\end{align*}

Here $\phi$ is learned online using an inverse dynamics model.
 The algorithms above include all possible combinations of global bonuses (second term) with the elliptical bonus (first term), and combining the two by multiplication or by taking a weighted sum. For the algorithms which take a weighted sum, we tuned the $\beta$ term on a subset of tasks, and report the best value on all $16$ tasks. 

\section{Additional Experiment Results}

\subsection{Additional MiniHack Results}
\label{appendix:extra-minihack-envs}

In Table \ref{tab:extra-minihack-envs} we report results for two additional MiniHack variants of the \texttt{MultiRoom} environment. The first features locked doors which the agent must kick down in order to move to the next room and eventually reach the exit. The second features moving monsters which the agent must fight or avoid while navigating towards the exit. In both cases, we see similar phenomena as in Section \ref{sec:episodic-advantages}: the global bonus exhibits a severe performance drop when going from $|\mathcal{C}|=1$ to $|\mathcal{C}|=\infty$, whereas the episodic bonus' performance is largely unchanged. 

\begin{table}[h!]
    \centering
    \begin{tabular}{c|c|c|c|c}
         Environment & $|\mathcal{C}|$ & $\psi$ & Global & Episodic \\
    \hline 
    \hline         
         \texttt{MultiRoom-N4-Locked-v0} & $1$ & P & $0.722 \pm 0.446$ & $0.948 \pm 0.01$ \\
         \texttt{MultiRoom-N4-Locked-v0} & $\infty$ & P & $-0.230 \pm 0.039$ & $0.925 \pm 0.01$ \\ 
         \texttt{MultiRoom-N6-LavaMonsters-v0} & $1$ & P & $0.994 \pm 0.001$ & $0.76 \pm 0.38$ \\
         \texttt{MultiRoom-N6-LavaMonsters-v0} & $\infty$ & P & $0.00 \pm 0.00$ & $0.75 \pm 0.21$ \\          
    \end{tabular}
    \caption{Reward for global and episodic bonuses for different CMDPs, averaged across $5$ seeds. Here $|\mathcal{C}|$ denotes the number of different contexts/maps which are sampled from at each episode. The $\psi$ column indicates which feature encodings are used (P for positions, M for messages).}
    \label{tab:extra-minihack-envs}
\end{table}

\subsection{Evolution of Global and Episodic bonuses}
\label{appendix:combination-evolution-results}
\begin{figure}[h]
    \centering
    \includegraphics[width=\textwidth]{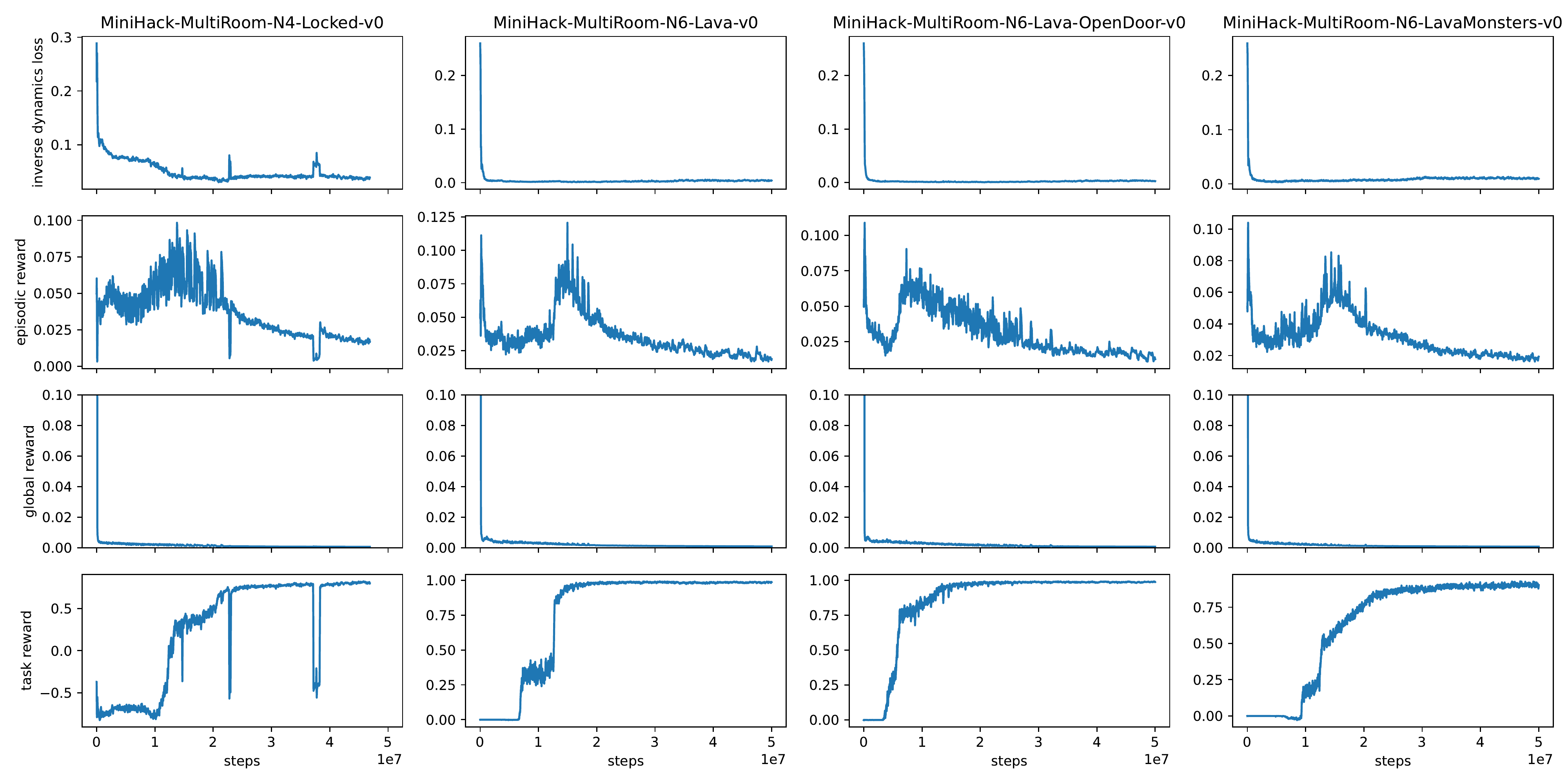}
    \caption{Evolution of inverse dynamics loss, episodic bonus, global bonus and extrinsic reward throughout training for \textsc{E3BxNovelD}.}
    \label{fig:bonus-evolution}
\end{figure}

In this section we show the evolution of the global and episodic bonus terms for the \textsc{E3BxNovelD} algorithm for four of the MiniHack tasks (see Figure \ref{fig:bonus-evolution}). The first row displays the inverse dynamics model loss used for learning the $\phi$ embedding in E3B's episodic bonus, the second row shows the E3B episodic bonus itself, the third shows the NovelD global bonus, and the fourth shows the true extrinsic reward provided by the environment. 

First, note that the global bonus spans a much larger range of values than the episodic bonus does, initially starting at a high value, exhibiting a first rapid decay, and then further decaying at a slower rate. In contrast, the episodic bonus spans a more limited range, and during most of the training it has higher magnitude than the global bonus. For the episodic bonus, we first see an initial decrease in magnitude, which is likely due to the $\phi$ features stabilizing (note that this coincides with the stabilization of the inverse dynamics loss used for learning $\phi$). After this, the episodic bonus increases, indicating that the agent's policy is learning to maximize the episodic bonus. The episodic bonus then decreases again once the extrinsic reward starts to increase, indicating that the agent has found the true environment reward and is switching to exploitation rather than exploration.

We hypothesize that additively combining the episodic and global bonuses does not offer much benefit over the episodic bonus alone because the magnitude of the global bonus decays significantly over time. Recall that the global bonus is computed using \textit{all} the agent's experience, so it eventually becomes exhausted. Since the episodic bonus is reset each episode, it does not become exhausted the same way, as evidenced by the fact that it \textit{increases} once the feature encoder $\phi$ has stabilized. If we add the two together, eventually the contribution of the global bonus will be small compared to the contribution of the episodic bonus. However, if we combine the two multiplicatively the global bonus will still have an effect regardless of its scale. 

\subsection{Results with NGU episodic bonus}
\label{appendix:ngu-results}

In this section we report results for the KNN-based episodic bonus from the Never-Give-Up (NGU) agent proposed in \citep{NGU}. Like the E3B bonus, the NGU bonus operates in the embedding space induced by an inverse dynamics model. A key difference is that it is based on the euclidean norm between the current state's embedding and its neareset neighbors within the episode, whereas E3B's bonus is computed using the metric induced by the episodic covariance matrix. 

\begin{figure}
    \centering
    \includegraphics[width=\textwidth]{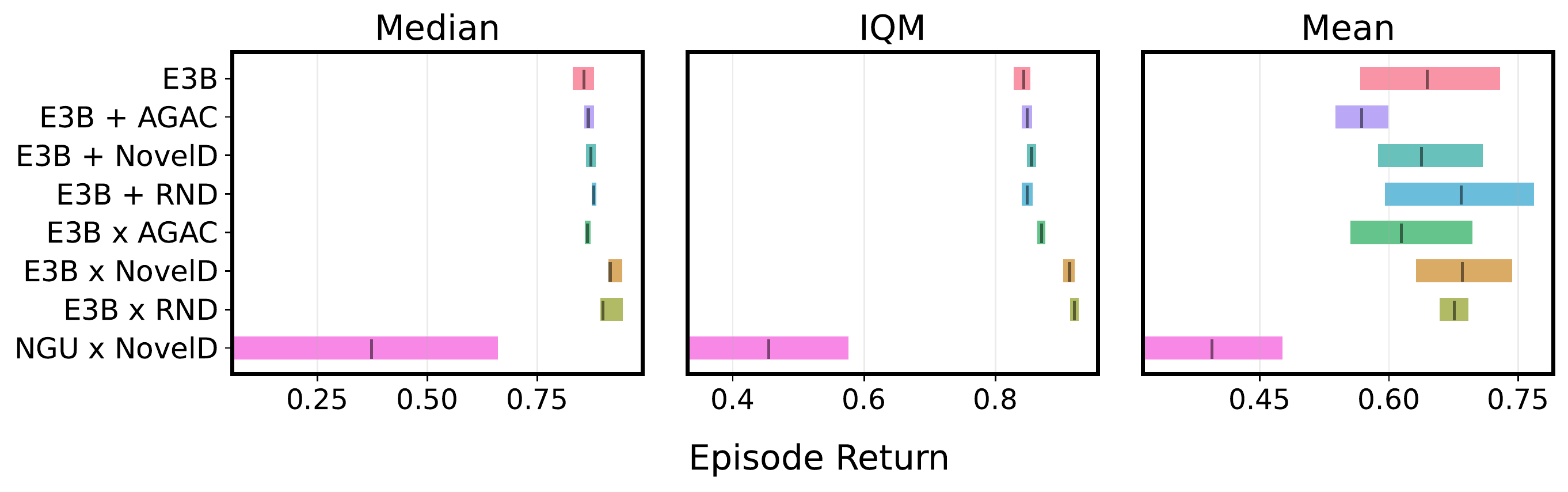}
    \caption{Performance for all methods including using NGU's episodic bonus.}
    \label{fig:aggregate-ngu}
\end{figure}

Results for a combination of NGU's episodic bonus with NovelD's global bonus are shown in Figure \ref{fig:aggregate-ngu}. Although this approach performs better than not including an exploration bonus (which gives average return of $0$), it performs considerably worse than the variants which use E3B. Note that our results are consistent with \citep{wang2023revisiting}, who also report poor results for NGU's episodic bonus on MiniGrid. 

One possible explanation may be that since NGU's KNN-based bonus uses the euclidean norm, if one dimension has higher scale than others it may dominate and reduce the effect of other more informative features. Therefore, this bonus may be more sensitive to spurious features in embedding space. On the other hand, E3B automatically adjusts the scale of each feature by normalizing by the inverse covariance matrix, and may therefore be more robust.

\clearpage

\section{Additional Discussion}

In this section we provide additional discussion and examples for the framework described in Section \ref{sec:framework}. 

\subsection{Examples of $\psi$ functions and $Z$ spaces}
\label{appendix:value-function-examples}

\begin{enumerate}
    \item \textbf{Tabular Algorithms:} In classical tabular algorithms with a UCB-style bonus \citep{E3, Rmax, MBIE, q-learning-provable}, $\psi$ is simply the identity and $Z = \mathcal{S}$. The bonus is of the form $b(s) = 1/\sqrt{N(\psi(s))} = 1/\sqrt{N(s)}$.
    
    \item \textbf{Deep RL Algorithms with position-based counts:} A number of recent exploration algorithms have used inverse counts of $(x, y)$ locations to drive exploration in gridworlds, for example RIDE \citep{RIDE, minihack}, AGAC \citep{AGAC} and NovelD \citep{NovelD, E3B}. We use this approach for our experiments on \texttt{MiniHack-MultiRoom} and \texttt{MiniHack-Corridor} in Section \ref{sec:motivation}. In this case, $\psi(s_t) = (x_t, y_t)$, the coordinates of the agent at time $t$, and the bonus is defined as $b(s) = 1/\sqrt{N(\psi(s_t))}$. The space $Z$ is a 2D lattice the size of the map, i.e. 
    \begin{equation*}
        Z=\{(i, j): 1 \leq i \leq W, 1 \leq j \leq H\}
    \end{equation*} 
    where $W, H$ correspond to the maximum width and height of the map across all contexts.
    
    \item \textbf{Deep RL Algorithms with message-based counts:} Some recent methods have used inverse counts based on textual messages \cite{mu2022improving, E3B} for environments where these are available, such as MiniGrid or MiniHack. We use this approach for the \texttt{MiniHack-KeyRoom} environment shown in Figure \ref{fig:keyroom-minihack}. In this case, $\psi$ extracts the message from the state and the bonus is computed as $b(s) = 1/\sqrt{N(\psi(s))}$. The space $Z$ then consists of all possible messages for the given environment. 
    For example, for the \texttt{MiniHack-KeyRoom} environment, we have 
    \begin{align*}
        Z = \{& \texttt{"You can't move diagonally out of an intact doorway."}, \\
& \texttt{"Be careful!  New moon tonight."}, \\
& \texttt{" "}, \\
& \texttt{"It's a wall."}, \\
& \texttt{"You see here a key named The Master Key of Thievery."}, \\
& \texttt{"h - a key named The Master Key of Thievery."}, \\
& \texttt{"The stairs are solidly fixed to the floor."}, \\
& \texttt{"It won't come off the hinges."}, \\
& \texttt{"You can't move diagonally into an intact doorway."}, \\
& \texttt{"This door is locked."}, \\
& \texttt{"Never mind."}, \\
& \texttt{"There is nothing here to pick up."} \\
\}
    \end{align*}
    
    \item \textbf{Elliptical Global Bonuses with kernel functions} The examples above have used count-based bonuses in feature space. However, our framework also captures algorithms which use exploration bonuses other than counts, such as elliptical bonuses. For example, the work of \cite{PCPG} uses an elliptical bonus in the space induced by an RBF kernel. In this case, $\psi$ is the mapping to kernel space, $Z = \mathbb{R}^n$ (where $n$ is the number of points used to compute the RBF kernel) and the bonus is given by $b(s_t) = \psi(s_t)^\top C_{t-1}^{-1} \psi(s_t)$. Here $C_{t-1} = \sum_{i=1}^{t-1}\psi(s_i)\psi(s_i)^\top$ is the (unnormalized) covariance matrix computed using all the agent's experience. 
    
    \item \textbf{Elliptical Episodic Bonuses with learned embedding} The recent E3B algorithm \cite{E3B} can be described within our framework as well. Here $\psi: \mathcal{S} \rightarrow Z$ is learned using an inverse dynamics model, and $Z = \mathbb{R}^k$ is the embedding space. The bonus is given by $b(s) = \psi(s_t)^\top C_t^{-1} \psi(s_t)$, where $C_{t-1} = \sum_{i=t_0}^{t-1}\psi(s_i)\psi(s_i)^\top$ is the episodic covariance matrix (with $t_0$ denoting the start of the current episode). Here $\psi$ is learned using an inverse dynamics model \citep{pathak2017curiosity}. 
        
\end{enumerate}

\subsection{Visualizations of $V^\star_{\psi, c}$ functions}
\label{appendix:value-function-viz}

\begin{figure*}
        \centering
        \begin{subfigure}[b]{0.32\textwidth}
            \centering
            \includegraphics[width=\textwidth]{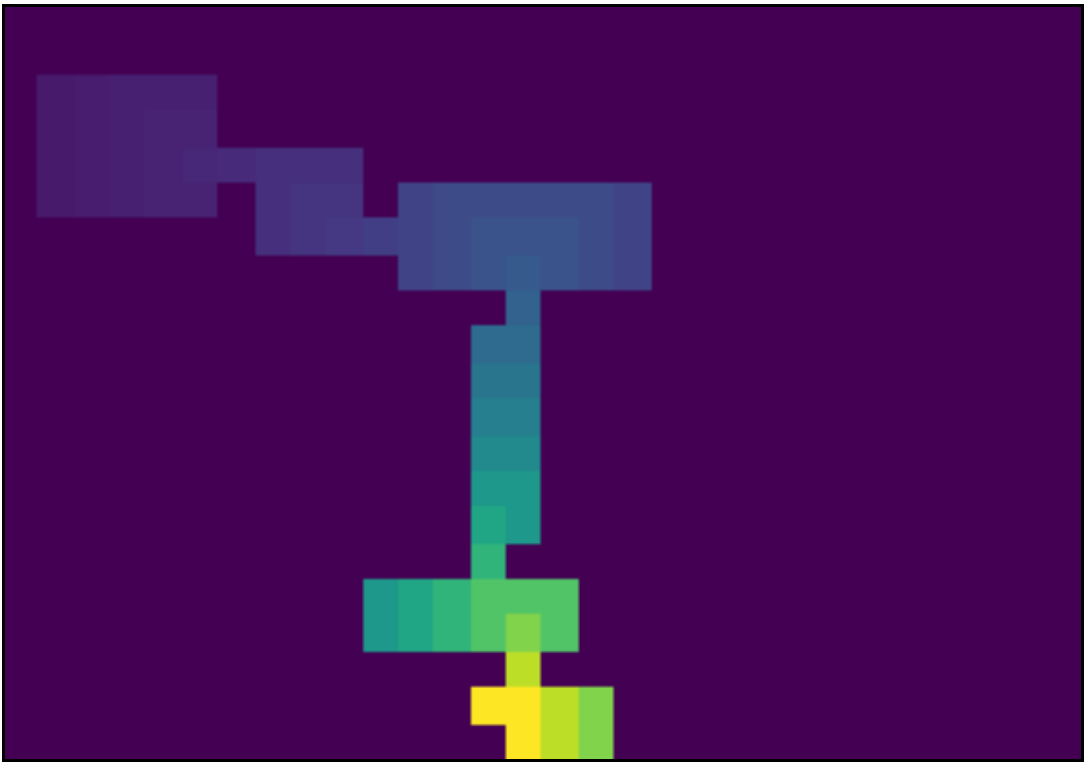}
            \label{fig:mean and std of net14}
        \end{subfigure}
        \hfill
        \begin{subfigure}[b]{0.32\textwidth}
            \centering
            \includegraphics[width=\textwidth]{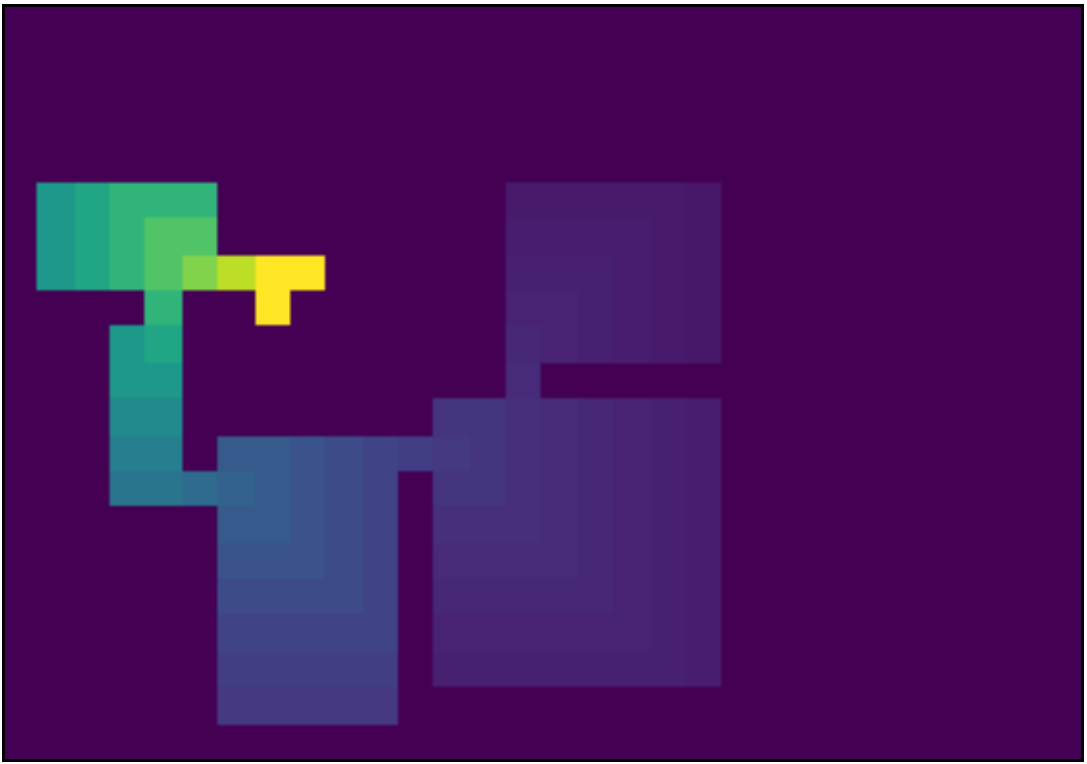}
            \label{fig:mean and std of net14}
        \end{subfigure}
        \hfill        
        \begin{subfigure}[b]{0.32\textwidth}  
            \centering 
            \includegraphics[width=\textwidth]{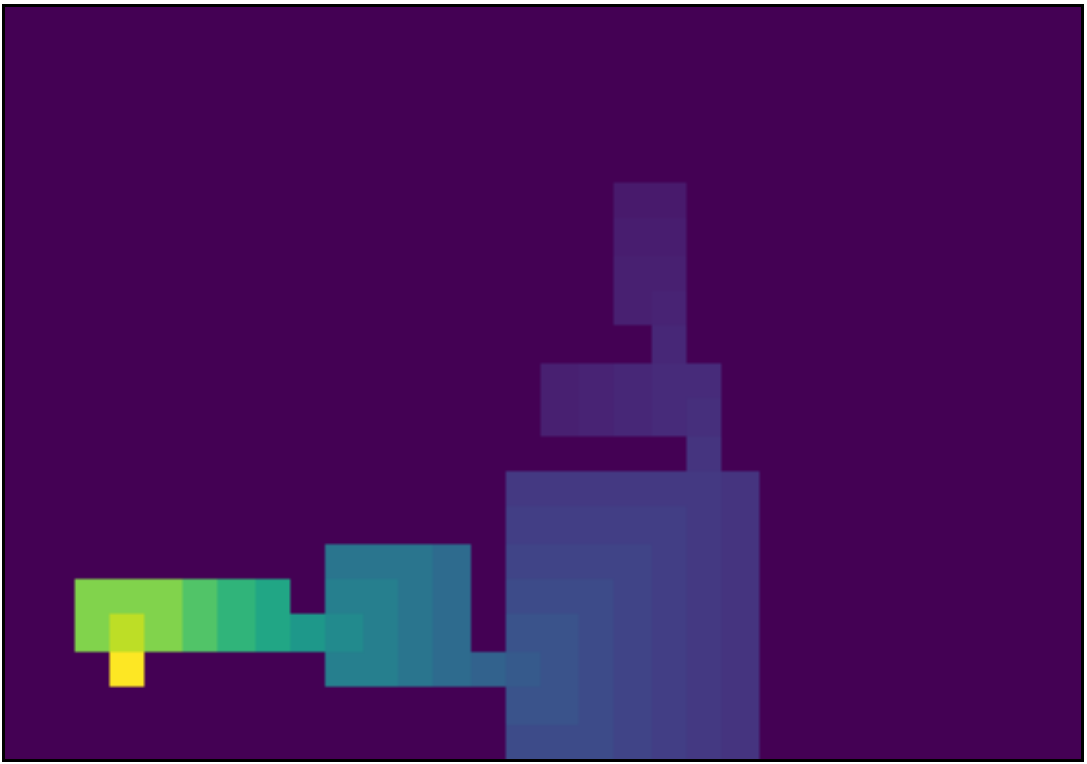}
            \label{fig:mean and std of net24}
        \end{subfigure}
        \vskip\baselineskip
        \begin{subfigure}[b]{0.32\textwidth}   
            \centering 
            \includegraphics[width=\textwidth, height=0.7\textwidth]{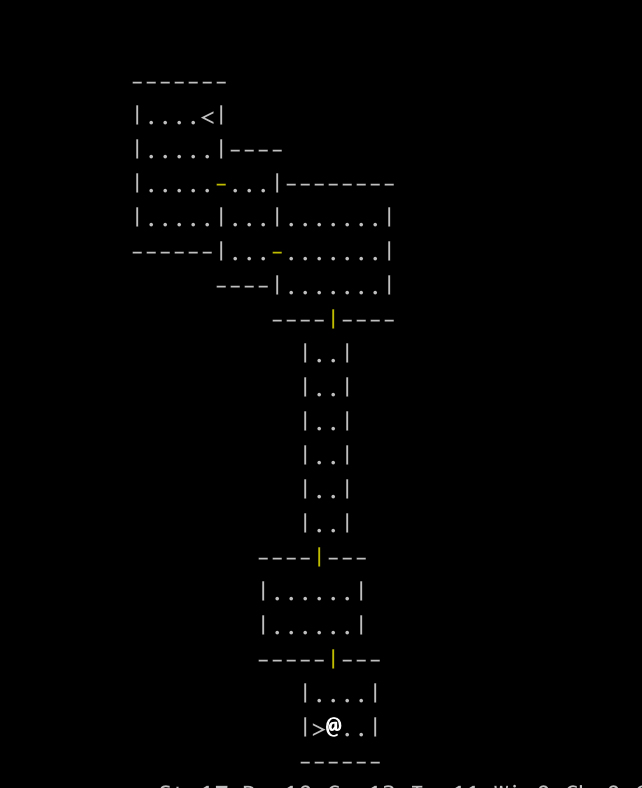}
            \caption[]%
            {$c=1$}    
            \label{fig:mean and std of net34}
        \end{subfigure}
        \hfill
        \begin{subfigure}[b]{0.32\textwidth}   
            \centering 
            \includegraphics[width=\textwidth, height=0.7\textwidth]{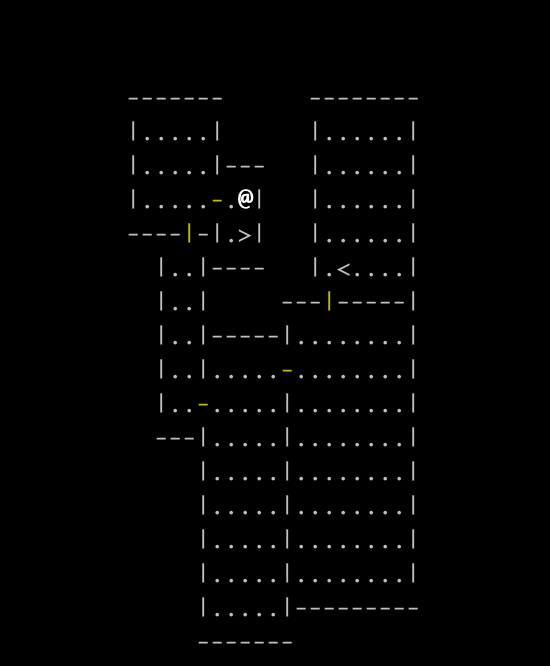}
            \caption[]%
            {$c=2$}    
            \label{fig:mean and std of net34}
        \end{subfigure}
        \hfill        
        \begin{subfigure}[b]{0.32\textwidth}   
            \centering 
            \includegraphics[width=\textwidth, height=0.7\textwidth]{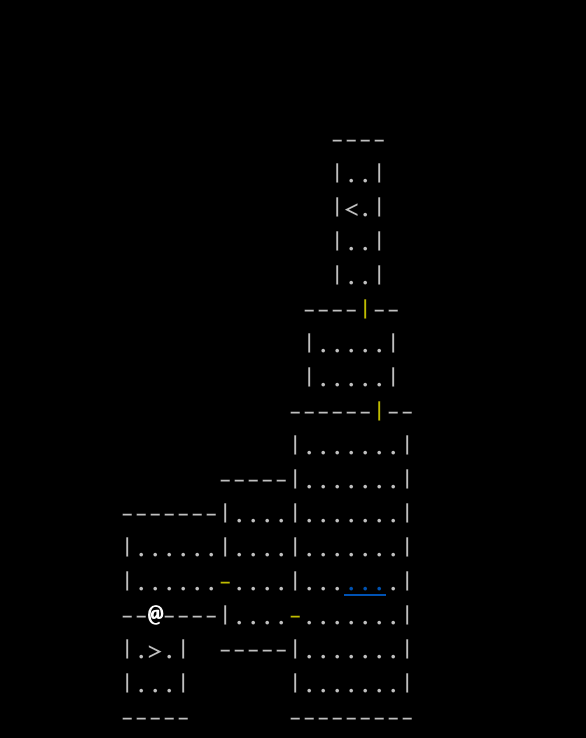}
            \caption[]%
            {$c=3$}    
            \label{fig:mean and std of net44}
        \end{subfigure}
        \caption{Top row: Value maps $V^\star_{\psi, c}$ for $3$ different contexts of the \texttt{MultiRoom} environment. For each $z=(x, y)$ location on the map, we display $V^\star_{\psi, c}(z) = \gamma^{k(x, y)}$, where $k(x, y)$ is the shortest path from the $(x, y)$ location to the goal and $\gamma=0.9$. Bottom row shows corresponding maps. Note that $V^\star_{\psi, c}$ changes significantly with different values of $c$.} 
        \label{fig:value-viz-multiroom}
    \end{figure*}

We now provide visualizations of the $V^\star_{\psi, c}$ functions defined in Examples 1 and 2 of Section \ref{sec:framework}.
Figure \ref{fig:value-viz-multiroom} shows the functions $V^\star_{\psi, c}$ for three different contexts $c$ in Example 1, corresponding to different maps, with $\psi: \mathcal{S} \rightarrow Z$ given by $\psi(s_t) = (x_t, y_t)$. Note that here $Z$ is a 2D lattice consisting of all possible $(x, y)$ positions on the map. For each $z=(x, y)$, we have $V^\star_{\psi, c}(z) = \gamma^{k(x, y)}$, where $k(x, y)$ denotes the shortest path to the goal computed using graph search. We use $\gamma=0.9$ for these visualizations. The function $V^\star_{\psi, c}$ changes significantly for different contexts (maps) $c$, since the geometry of the map and the goal location change from one context to the other.  

\begin{figure}
     \centering
     \includegraphics{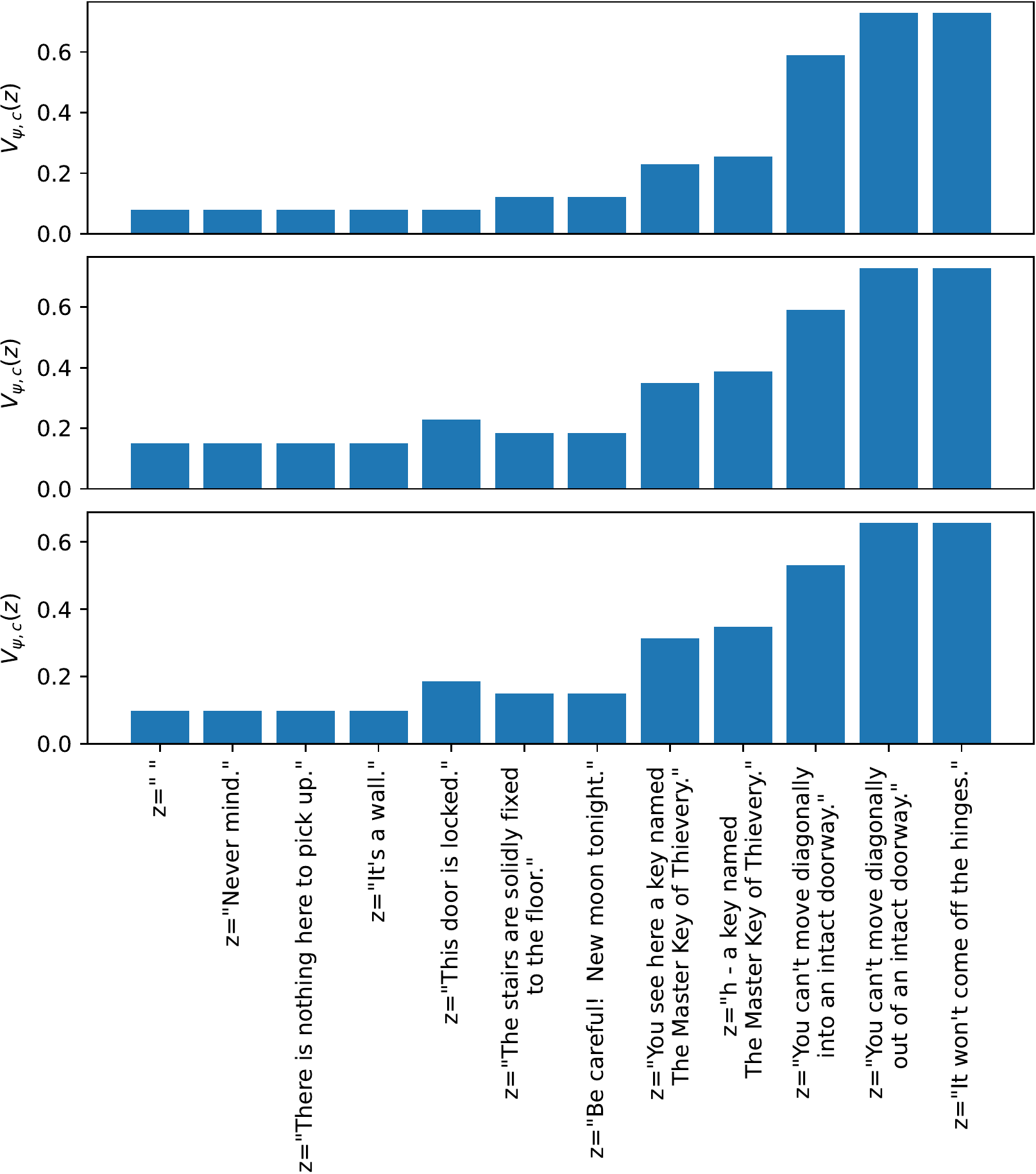}
     \caption{Visualization of value maps $V^\star_{\psi, c}$ for $3$ different contexts for the \texttt{KeyRoom} environment where $\psi$ encodes messages. Here $V^\star_{\psi, c}$ changes little with different contexts.}
     \label{fig:value-viz-keyroom}
 \end{figure}

Next, in Figure \ref{fig:value-viz-keyroom} we visualize the $V^\star_{\psi, c}$ function for the \texttt{MiniHack-KeyRoom} environment (shown in Figure \ref{fig:keyroom-minihack}), with $\psi: \mathcal{S} \rightarrow Z$ extracting messages from states. In this case $Z$ consists of the set of all possible messages which can be seen in this environment. We show $V^\star_{\psi, c}$ for $3$ different contexts $c$. Unlike in the previous example, here $V^\star_{\psi, c}$ has similar shape across all contexts. Messages which can only occur once the door has opened, such as \texttt{"It won't come off the hinges"} and \texttt{"You can't move diagonally out of an intact doorway"}, have the highest value. Messages corresponding to the agent visiting or picking up the key, which is necessary for opening the door, have medium value (e.g. \texttt{"You see here a key names The Master Key of Thievery"}, \texttt{"h - a key named The Master Key of Thievery"}. The remaining messages, such as \texttt{" "}, \texttt{"This door is locked"}, \texttt{"It's a wall"}, do not indicate that the agent has made progress on the task and have the lowest value.

  \begin{figure}
     \centering
     \includegraphics{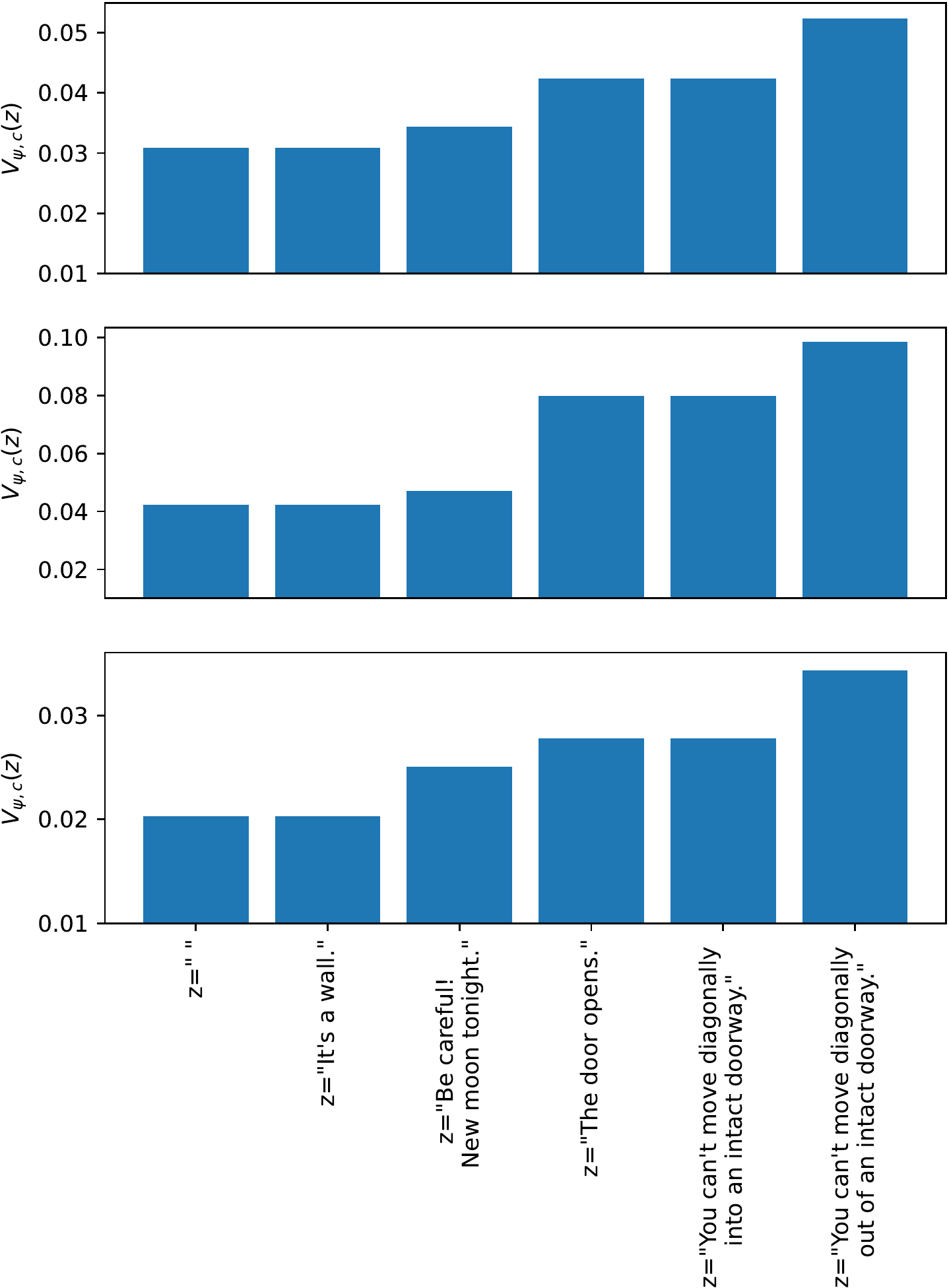}
     \caption{Visualization of value maps $V^\star_{\psi, c}$ for $3$ different contexts for the \texttt{MultiRoom} environment where $\psi$ encodes messages. Here $V^\star_{\psi, c}$ changes little with different contexts.}
     \label{fig:value-viz-multiroom-msg}
 \end{figure}

\clearpage

\subsection{Quantifying the Variation of $V^\star_{\psi, c}$ across episodes}
\label{appendix:quantifying-variation}

One way in which the variation of $V^\star_{\psi, c}$ across contexts can be made precise is to take the average cosine similarity between $V^\star_{\psi, c}$ and $V^\star_{\psi, c'}$ for randomly sampled contexts $c, c'$. Note that this is well-defined if $Z$ is finite or infinite---if $Z$ is infinite, we replace dot products between vectors by inner products between functions:

 \begin{equation}
     \mbox{average cosine similarity} = \mathbb{E}_{c, c' \sim \mu_C}\Big[ \frac{\langle V^\star_{\psi, c}, V^\star_{\psi, c'} \rangle}{\|V^\star_{\psi, c}\| \cdot \|V^\star_{\psi, c'}\|}\Big]
 \end{equation}




Average cosine similarities for all three examples in Section \ref{sec:framework} are given in Table \ref{tab:avg-similarity-examples}. 
This is consistent with what we observe in the visualizations in Appendix \ref{appendix:value-function-viz}, where the $V^\star_{\psi, c}$
 for Example 1 appear very different across contexts whereas those for Examples 2 and 3 appear very similar. It is also consistent with our experimental results where the episodic bonus succeeds for Example 1 and the global bonus succeeds for Examples 2 and 3.

\begin{table}[]
    \centering
    \begin{tabular}{c|l|l|c}
         Example & Environment & $\psi$ & Average Cosine Similarity  \\
         \hline
         \hline
         1 & MultiRoom & positions & $0.198$ \\
         2 & KeyRoom & messages & $0.991$ \\
         3 & MultiRoom & messages & $0.951$
    \end{tabular}
    \caption{Average cosine similarity for different examples from Section \ref{sec:framework}.}
    \label{tab:avg-similarity-examples}
\end{table}

It is difficult to say in general at what exact rate the global performance will degrade with decreasing average cosine similarity of $V^\star_{\psi, c}$ across contexts. However, to get some idea we can check this empirically for our \texttt{MultiRoom} experiments with position encodings and different $|\mathcal{C}|$ from Table \ref{tab:count-results}. In Table \ref{tab:cosine-sim-vs-perf} below we compute the average cosine similarity of $V^\star_{\psi, c}$ across contexts and compare this to the global bonus performance. Again, we see that the global bonus performs worse in settings with low average cosine similarity. 

\begin{table}[h!]
    \centering
    \begin{tabular}{c|c|c|c|c}
        Environment & $\psi$ & $|\mathcal{C}|$ & Average Cosine Similarity & Global Bonus Performance  \\
        \hline
        \hline
        MultiRoom & position & $1$ & $1.000$ & $0.99$ \\
        MultiRoom & position & $3$ & $0.465$ & $0.59$ \\
        MultiRoom & position & $5$ & $0.358$ & $0.23$ \\
        MultiRoom & position & $10$ & $0.278$ & $0.02$ \\
        MultiRoom & position & $\infty$ & $0.198$ & $0.00$         
    \end{tabular}
    \caption{Average cosine similarity vs. global bonus performance on MultiRoom for different numbers of contexts.}
    \label{tab:cosine-sim-vs-perf}
\end{table}

\clearpage

\end{document}